\documentclass[iicol,pdflatex,sn-mathphys-num]{sn-jnl}
\graphicspath{{figures/}}

\usepackage{graphicx}
\usepackage{amsmath,amssymb,amsfonts}
\usepackage{textcomp}
\usepackage{xcolor}
\usepackage{booktabs}
\usepackage{algorithm}
\usepackage{algpseudocode}
\usepackage{microtype}
\usepackage[htt]{hyphenat}
\usepackage{placeins}
\usepackage{needspace}
\hypersetup{hidelinks}
\makeatletter
\def\ps@headings{%
  \let\@mkboth\markboth
  \def\@oddfoot{\hfill\thepage\hfill}%
  \let\@evenfoot\@oddfoot
  \def\@evenhead{%
    \vbox to 0pt{\vspace*{-48pt}%
      \hbox to \hsize{\footnotesize\itshape Jerry Adams Franklin\hfill}}\par
    \hspace*{-\textwidth}\hbox to \hsize{\hfill}}%
  \def\@oddhead{%
    \vbox to 0pt{\vspace*{-48pt}%
      \hbox to \hsize{\hfill\footnotesize\itshape\leftmark}}\par
    \hspace*{-\textwidth}\hbox to \hsize{\hfill}}%
}
\def\ps@firstpage{%
  \let\@mkboth\markboth
  \def\@oddfoot{\hfill\thepage\hfill}%
  \let\@evenfoot\@oddfoot
  \def\@oddhead{}\def\@evenhead{}}
\def\email#1{\global\advance\emailcnt by 1\relax
  \if@corauemail
    \g@addto@macro\corrauthemail{%
      \setcounter{footnote}{0}%
      \textcolor{black}{#1}%
    }%
  \else
    \g@addto@macro\authemail{%
      \setcounter{footnote}{0}%
      \textcolor{black}{#1}%
    }%
  \fi}
\makeatother
\providecommand{\doiurl}[1]{\url{https://doi.org/#1}}

\begin{document}

\title[Measured Comm.\ Frontier for Federated LoRA]{A Measured Communication-Quality Frontier for Federated LoRA Fine-Tuning with Adaptive Phase-Switching}

\author*{\fnm{Jerry Adams} \sur{Franklin}}\email{jerry.adamsf@gmail.com}
\affil*{\orgname{Independent Researcher}, \orgaddress{\city{Fairport}, \state{NY}, \postcode{14450}, \country{USA}}}

\abstract{Federated fine-tuning of large language models with low-rank adaptation (LoRA) reduces the number of trainable parameters, but communication remains the dominant cost, and protocols are usually compared by parameter-count ratios rather than by measured bytes. This paper measures per-round upload and download bytes for five federated LoRA protocols, three of them from prior work, and places them on a single communication-quality frontier scored by held-out instruction-following loss. The frontier has a knee. ReverseAdaptive, which learns both LoRA factors before freezing one once the relative improvement in training loss falls below a dimensionless threshold, sits at that knee: it cuts measured round-trip communication by 40.5\% relative to FLoRA at a held-out loss cost of 0.0063, and beats FFA-LoRA, which freezes that factor at initialization, by 0.0182 in held-out loss, more than twenty times the largest per-method seed standard deviation. The same threshold carries to LLaMA-3.2-3B without retuning, where it saves 30.0\%.}


\maketitle
{\renewcommand\thefootnote{}\footnotetext{Preprint of a manuscript submitted for journal publication. Code, configurations, and run records are archived at \url{https://doi.org/10.5281/zenodo.23111915}.}}
\pagestyle{headings}
\thispagestyle{firstpage}

\section{Introduction}
\label{sec:intro}

Federated learning enables model training across distributed clients without sharing raw data, making it attractive for privacy-sensitive applications of large language models (LLMs) \cite{kairouz2021advances}. Fine-tuning with low-rank adaptation (LoRA) \cite{hu2022lora} substantially reduces the number of trainable parameters per client, but federated LoRA aggregation still transmits nontrivial bytes per round. Because adapters cross the network in every round, cumulative communication often determines whether a protocol is deployable where uplink bandwidth is scarce or metered.

Existing federated LoRA methods differ in what they aggregate and when. FedIT \cite{zhang2024federatedgpt} applies FedAvg \cite{mcmahan2017communication} directly to LoRA's $A$ and $B$ matrices, paying full bidirectional cost per round. FFA-LoRA \cite{sun2024ffalora} freezes $A$ at its initialization so that clients train and transmit only $B$ from the first round, giving a theoretical 50\% upload saving. FLoRA \cite{wang2024flora} stacks client LoRA modules and compresses via singular value decomposition (SVD), preserving product-space semantics at full bidirectional cost. FlexLoRA \cite{bai2024flexlora} supports heterogeneous client ranks and is orthogonal to the communication protocol studied here. These four report communication savings as parameter-count ratios rather than measured bytes; more recent work does measure transmitted volume directly \cite{yan2026fedsrd,ramesh2026florist}. To the author's knowledge, no study has placed these protocols on one measured communication axis against a quality metric that separates them. Nor, to the author's knowledge, does any study report the asymmetric round that a mid-training change of aggregation mode creates.

This paper addresses these gaps with three contributions.

\textbf{A five-method frontier and its knee.}
Five protocols, three from prior work, are placed on one measured frontier on Alpaca at TinyLlama-1.1B scale; three of them are also compared on Dolly-15k and at LLaMA-3.2-3B. Quality is evaluated by held-out instruction-following loss rather than by zero-shot benchmarks, which do not discriminate between protocols at the smaller scale; at LLaMA-3.2-3B, where held-out loss was not computed, final training loss and downstream accuracy are reported. The frontier has a knee at ReverseAdaptive: savings beyond that point cost roughly five times more quality per point. Total communication follows a closed form in the switch round that predicts every measured total at full participation to the byte.

\textbf{Bidirectional B-only protocol with measured byte tracking.}
The protocol implements both upload and download B-only transmission for Two-Phase and ReverseAdaptive aggregators. Per-round upload and download megabytes are recorded from the tensors actually transmitted. The transition round, in which the server seeds its frozen $A$ from client uploads, is accounted for explicitly, producing a one-round asymmetry between when download savings begin and when upload savings begin. Two internal checks bound the instrument: FedIT, implemented independently as a full-state protocol, lands on FLoRA's measured total to the byte, and disabling the switch reproduces the matched FLoRA run to 10 decimal places.

\textbf{ReverseAdaptive: adaptive phase-switching aggregator.}
The aggregator starts in FLoRA mode and monitors the relative per-round improvement in global training loss. When that relative improvement falls below a dimensionless threshold $\tau$ for the first time after a warmup period, the aggregator switches to FFA-LoRA permanently. Varying $\tau$ traverses the same communication-quality frontier as the fixed-$K$ Two-Phase family without requiring $K$ to be selected in advance. The comparison against FFA-LoRA measures the effect of learning $A$ before freezing it rather than freezing it at initialization.

\textbf{Summary of results.}
Two-Phase $K{=}8$ saves 27.7\% (TinyLlama-1.1B) and 26.0\% (LLaMA-3.2-3B) over FLoRA. ReverseAdaptive saves 40.5\% and $30.0 \pm 4.0$\% respectively, and FFA-LoRA saves 61.9\% at TinyLlama scale for 0.0182 more held-out loss than ReverseAdaptive. At LLaMA-3.2-3B the same $\tau$ switches at rounds 7 to 9 without retuning.

Code, configuration files, and scripts to reproduce every figure and table are publicly released; see Appendix~\ref{app:repro}.

\needspace{5\baselineskip}
\section{Related Work}
\label{sec:related}

\subsection{LoRA and Parameter-Efficient Fine-Tuning}
Parameter-efficient fine-tuning (PEFT) methods reduce the cost of adapting large pretrained models by training only a small subset of parameters. LoRA \cite{hu2022lora} decomposes weight updates into low-rank products $BA$, where $B$ is $d \times r$ and $A$ is $r \times k$ with $r$ much smaller than $\min(d,k)$. This reduces the number of trainable parameters substantially, enabling fine-tuning of large models without modifying base weights. The resulting LoRA adapters are compact and composable, making them a natural fit for federated settings where both storage and communication are constrained.

\subsection{Federated LoRA Aggregation}
FedIT \cite{zhang2024federatedgpt} applies standard FedAvg \cite{mcmahan2017communication} to LoRA matrices: the global $A$ and $B$ are computed as weighted averages of client $A$ and $B$ matrices. The aggregation introduces a mismatch: the optimal global update is the product $BA$, not a product of separately averaged $B$ and $A$. Averaging independently introduces aggregation error that grows with client heterogeneity.

FFA-LoRA \cite{sun2024ffalora} addresses the aggregation mismatch by freezing $A$ at initialization and training only $B$ across clients. Because $A$ is constant, the server aggregates $B$ matrices directly without product-space distortion. The theoretical upload saving is 50\%. The original work also motivates the frozen-$A$ design in terms of differential privacy. A critical difference from the protocol in this paper: FFA-LoRA freezes $A$ at initialization, whereas the bidirectional B-only protocol allows $A$ to be learned through a FLoRA phase before freezing, recovering expressivity at the cost of longer initial overhead.

FLoRA \cite{wang2024flora} preserves product-space semantics by stacking client LoRA modules horizontally and applying truncated SVD to recover a global low-rank adapter, reducing the aggregation bias present in FedIT. The tradeoff is communication cost: FLoRA transmits full $A{+}B$ payloads in both directions every round. FLoRA is the full-state baseline in the experiments below. FedSA-LoRA \cite{guo2025fedsalora} makes the opposite choice, sharing only $A$ on the grounds that $A$ carries general knowledge while $B$ is client-specific; under the byte accounting of Section~\ref{sec:implementation}, sharing $A$ instead of $B$ transmits 64\% of the full TinyLlama state rather than 36\%.

FlexLoRA \cite{bai2024flexlora} extends federated LoRA to the heterogeneous-rank setting, where different clients use different LoRA ranks based on local compute budgets. FlexLoRA's contribution is orthogonal to the protocol studied here; the bidirectional B-only protocol applies equally to homogeneous-rank clients.

\subsection{Communication-Efficient Federated Learning}
Communication efficiency in federated learning has been a central concern since FedAvg \cite{mcmahan2017communication} introduced local multi-step updates to reduce communication rounds. The broader challenge of open problems in federated learning is surveyed by Kairouz et al. \cite{kairouz2021advances}. Subsequent work has pursued per-round payload reduction through gradient compression. QSGD \cite{alistarh2017qsgd} reduces the bit-width of transmitted gradients through stochastic quantization. Deep gradient compression \cite{lin2017deep} applies aggressive sparsification, sending only 0.1\% of gradients per round with error feedback. FedPAQ \cite{reisizadeh2020fedpaq} combines periodic averaging with quantization, showing that 8-bit transmission incurs negligible accuracy loss. These techniques are orthogonal to the B-only protocol studied here and could in principle be composed with it for additional savings.

\subsection{Adaptive Aggregation and Scheduling}
Several federated learning methods adapt their behavior based on observed training dynamics. FedProx \cite{li2020fedprox} adds a proximal term to the local objective, controlling deviation from the global model. FedNova \cite{wang2020fednova} normalizes local gradients by effective step count, adaptively correcting for heterogeneous local training. FedAdam and FedYogi \cite{reddi2020adaptive} apply adaptive moment estimation at the server level, treating the aggregated pseudo-gradient as input to Adam-style updates. To the author's knowledge, no prior federated LoRA work uses a loss-plateau signal to switch between distinct aggregation modes during training.

\needspace{5\baselineskip}
\section{Method}
\label{sec:method}

\subsection{Background and Notation}
Each of $N$ clients holds a local dataset. A shared base model has weight matrices $W$. LoRA \cite{hu2022lora} parameterizes updates as $\Delta W = BA$ where $B$ is $d \times r$ and $A$ is $r \times k$. In each federated round, clients train their local adapters for one epoch, upload some representation to the server, and receive a global state.

The communication cost of a round is the sum of upload bytes and download bytes. The B-only fraction of the full state is not the 50\% that counting parameters alone suggests, because grouped-query attention (GQA) fixes the ratio of $B$ to $A$ parameters; Section~\ref{sec:implementation} derives the exact fractions of 36\% for TinyLlama-1.1B and 40\% for LLaMA-3.2-3B.

\subsection{Bidirectional B-Only Protocol}
\textbf{Phase 1 (FLoRA, rounds 1 to $K$).}
The server runs FLoRA aggregation \cite{wang2024flora}: client states are stacked and compressed via SVD to produce a global $(A_{\text{global}}, B_{\text{global}})$. Both upload and download payloads are full state.

\textbf{Transition (round $K{+}1$).}
Clients train with $A$ frozen at the broadcast round-$K$ global value and still upload full state $(A_i, B_i)$, because the server needs the $A$ matrices to compute and cache $A_{\text{frozen}}$. The server runs FFA-LoRA \cite{sun2024ffalora} aggregation for the first time, caches $A_{\text{frozen}}$, and broadcasts only $B_{\text{global}}$; clients keep the $A$ they already hold.

\textbf{Phase 2 steady state (rounds $K{+}2$ onward).}
Clients upload only $B_i$; the server reconstructs the full state as $(A_{\text{frozen}}, B_i)$ before aggregation. The server broadcasts only $B_{\text{global}}$; clients retain $A_{\text{frozen}}$ locally.

\textbf{Look-ahead download accounting.}
This produces a one-round asymmetry: download savings begin one round before upload savings. The experiment runner records this consistently, ensuring reported totals are accurate rather than optimistic.

Algorithm~\ref{alg:bonly} specifies the server-side decision logic per round.

\begin{algorithm}[t]
\centering
\caption{Bidirectional B-only protocol (server, per round)}
\label{alg:bonly}
\footnotesize
\begin{algorithmic}[1]
\Require round $r$, phase boundary $K$, frozen-$A$ cache (initially empty)
\Ensure upload mode, broadcast mode
\If{$r \leq K$} \Comment{FLoRA phase}
    \State upload $\leftarrow$ full state $(A, B)$
    \State broadcast $\leftarrow$ full state
\ElsIf{$r = K+1$} \Comment{transition: seed frozen $A$}
    \State upload $\leftarrow$ full state $(A, B)$
    \State $A_{\text{frozen}} \leftarrow$ $A$ from the first client's uploaded state
    \State broadcast $\leftarrow$ $B$ only \Comment{clients keep round-$K$ $A$}
\Else \Comment{FFA-LoRA steady state}
    \State upload $\leftarrow$ $B$ only \Comment{server prepends $A_{\text{frozen}}$}
    \State broadcast $\leftarrow$ $B$ only \Comment{clients keep $A_{\text{frozen}}$}
\EndIf
\end{algorithmic}
\end{algorithm}

\subsection{ReverseAdaptive Aggregator}
ReverseAdaptive replaces the fixed boundary $K$ with an adaptive loss-plateau signal. After a warmup period of $W$ rounds (default $W{=}5$), it monitors the \emph{relative} per-round loss improvement $\rho_r = (\ell_{r-1} - \ell_r) / \ell_{r-1}$. When $\rho_r < \tau$ for the first time, the aggregator switches to FFA-LoRA \cite{sun2024ffalora} mode permanently and instantaneously, with no intermediate phase. Because $\rho_r$ is a ratio of losses, $\tau$ is a dimensionless fraction rather than a loss difference, and its interpretation does not depend on the absolute scale of the loss for a given model or dataset; this scale-free construction is intended to let a single $\tau$ carry across model sizes without retuning, which Section~\ref{sec:choosing} examines. A complementary stability detector reverts to FLoRA if loss increases by more than 10\% after switching. No revert occurred in any ReverseAdaptive run reported in this paper, across both datasets, both model scales, all three partition settings, and the full threshold ablation.

The choice of $\tau$ trades communication savings against final training loss. Small $\tau$ delays the switch; large $\tau$ triggers it at the first eligible round after warmup. Section~\ref{sec:ablation} characterizes this tradeoff empirically.

The loss-plateau signal is chosen for its simplicity, its availability without additional instrumentation, and the fact that it tracks the quantity a practitioner optimizes. Alternative signals such as the gradient norm of $A$ or the singular value spectrum of stacked client matrices offer a different view of whether $A$ has converged and are left to future work.

Algorithm~\ref{alg:reverseadaptive} specifies the aggregator's per-round decision logic.

\begin{algorithm}[t]
\centering
\caption{ReverseAdaptive aggregator (per round)}
\label{alg:reverseadaptive}
\footnotesize
\begin{algorithmic}[1]
\Require round $r$, current loss $\ell_r$, prior loss $\ell_{r-1}$,
         warmup $W$, threshold $\tau$, current mode $m$, \texttt{client\_states}
\Ensure \texttt{global\_state}, updated mode $m$
\State $\rho_r \leftarrow (\ell_{r-1} - \ell_r) \,/\, \ell_{r-1}$ \Comment{relative improvement}
\If{$m = \text{FLoRA}$ \textbf{and} $r > W$ \textbf{and} $\rho_r < \tau$}
    \State $m \leftarrow \text{FFA-LoRA}$
    \State log event $(r, \texttt{switch\_to\_ffa})$
\ElsIf{$m = \text{FFA-LoRA}$ \textbf{and} $\ell_r > 1.1 \cdot \ell_{r-1}$} \Comment{revert}
    \State $m \leftarrow \text{FLoRA}$
    \State reset $A_{\text{frozen}}$
    \State log event $(r, \texttt{revert\_to\_flora})$
\EndIf
\If{$m = \text{FFA-LoRA}$}
    \State \Return \textsc{FFA-LoRA-Aggregate}(\texttt{client\_states})
\Else
    \State \Return \textsc{FLoRA-Aggregate}(\texttt{client\_states})
\EndIf
\end{algorithmic}
\end{algorithm}

\subsection{Implementation Details}
\label{sec:implementation}
\textbf{Byte tracking.}
Per-round upload and download megabytes are recorded in \texttt{results.json}. Megabytes are reported as $2^{20}$ bytes. Byte counts use the dtype in which parameters are transmitted: float32 (4 bytes/param) for TinyLlama-1.1B and float16 (2 bytes/param) for LLaMA-3.2-3B, whose LoRA parameters are trained in float32 and cast back to the base dtype for transmission. A full TinyLlama LoRA state is 2{,}252{,}800 parameters over 22 layers, or 9{,}011{,}200 bytes per client per direction. All models are optimized using AdamW \cite{loshchilov2019decoupled}.

\textbf{Frozen-$A$ provenance.} At the transition round the server seeds $A_{\text{frozen}}$ from the first client's uploaded state rather than from an average across clients. The two switching protocols reach that round differently. Two-Phase fixes its boundary in advance, so clients train round $K{+}1$ with $A$ already frozen at the broadcast round-$K$ global value; the first client's $A$ is exactly that value, and the server and every client hold the same $A$ for the rest of training. ReverseAdaptive decides to switch only after the switch round's local training, which runs with $A$ trainable, so its $A_{\text{frozen}}$ is the preceding round's global $A$ plus one epoch of the first client's local adaptation at learning rate $10^{-4}$. The broadcast that follows carries only $B$, so clients never receive this value and train $B$ for the remaining rounds against the $A$ they hold at the end of the switch round. In the simulation, which trains clients in sequence on one shared model, that is the last client's locally adapted $A$, while the first client's $A$ enters the aggregated and evaluated adapter. The two differ by one epoch of local adaptation on different client data, and the effect of this mismatch on ReverseAdaptive's reported quality is not measured. Client ordering is fixed for the duration of a run, so the behavior is deterministic and reproducible. Averaging $A$ across clients at the transition round is an alternative left to future work.

\textbf{GQA B-fraction.}
For TinyLlama-1.1B with rank $r{=}16$ targeting $\{\texttt{q\_proj}, \texttt{v\_proj}\}$, the projections adapted in the original LoRA formulation \cite{hu2022lora}, accounting for GQA in the value projection, the B-only fraction of the full state is exactly 36\% rather than the 50\% a naive parameter count would suggest. The same fraction holds for the four-projection attention set, since \texttt{q\_proj} and \texttt{o\_proj} share a shape and \texttt{k\_proj} and \texttt{v\_proj} share a shape under GQA, but not for target sets that break that pairing, such as \texttt{q\_proj} alone or sets including multilayer perceptron (MLP) projections. LLaMA-3.2-3B's different GQA configuration yields exactly 40\%.

\textbf{Partition mechanics.}
Alpaca carries no class labels, so partitions that are not independent and identically distributed (non-IID) apply Dirichlet skew over a task-diversity proxy, and the same proxy is applied to Dolly-15k rather than its category field so that both datasets are partitioned identically: each instruction $s$ is assigned to one of ten buckets by character length, $\min(\lfloor |s| / 50 \rfloor, 9)$, and a Dirichlet distribution with concentration $\alpha \in \{0.5, 0.1\}$ is applied over those buckets. This induces heterogeneity in task form rather than task semantics; Section~\ref{sec:limitations} notes that semantic heterogeneity is untested.

\section{Experiments and Results}
\label{sec:experiments}

\subsection{Setup}
\label{sec:setup}
Table~\ref{tab:setup} summarizes the experimental configuration. The corpus was executed on two hardware backends. A core of 34 runs on Apple silicon through the Metal Performance Shaders (MPS) backend, approximately 285 hours of compute, was re-executed on NVIDIA graphics processing units (GPUs), an RTX 4090 for TinyLlama-1.1B and an A100 40\,GB for LLaMA-3.2-3B; the cross-backend comparison is reported in Appendix~\ref{app:crossbackend}. The Dolly-15k replication and the FFA-LoRA and FedIT baseline runs were produced on the same rented hardware. Both TinyLlama-1.1B and LLaMA-3.2-3B experiments use three seeds for IID training.

\begin{table*}[tbp]
\centering
\caption{Experimental setup}
\label{tab:setup}
\setlength{\tabcolsep}{6pt}
\renewcommand{\arraystretch}{0.95}
\footnotesize
\begin{tabular}{@{}lp{0.72\textwidth}@{}}
\toprule
Parameter & Value \\
\midrule
Base models & TinyLlama-1.1B-Chat-v1.0 \cite{zhang2024tinyllama}; LLaMA-3.2-3B \cite{dubey2024llama3} \\
Datasets & Alpaca \cite{taori2023alpaca}; Dolly-15k \cite{dolly2023}; 3000 samples each \\
Data partitions & IID; Dirichlet $\alpha{=}0.5$; Dirichlet $\alpha{=}0.1$ \\
Clients / round & 10 / 10 (full participation) \\
Rounds & 15 \\
Local epochs & 1 \\
Batch size & 4 (TinyLlama), 2 (LLaMA-3.2-3B) \\
Grad.\ accumulation & 4 (TinyLlama), 8 (LLaMA-3.2-3B); effective batch 16 \\
Learning rate & $10^{-4}$ (AdamW \cite{loshchilov2019decoupled}) \\
Max sequence length & 256 \\
LoRA rank / alpha & 16 / 32 \\
LoRA target modules & \texttt{q\_proj}, \texttt{v\_proj} \\
LoRA dropout & 0.1 \\
Hardware (primary) & Apple M4 Pro, 48\,GB unified memory, MPS \\
Hardware (reproduction) & NVIDIA RTX 4090 (TinyLlama); A100 40\,GB (LLaMA-3.2-3B) \\
Training dtype & float32 (TinyLlama-1.1B); float16 base + float32 LoRA (LLaMA-3.2-3B) \\
Transmitted dtype & float32 (TinyLlama-1.1B); float16 (LLaMA-3.2-3B) \\
Downstream eval & 500 examples/benchmark, seed 42, log-likelihood \\
Held-out eval & 500 examples, train indices 3001--3500, seq.\ len.\ 256 \\
Benchmarks & MMLU \cite{hendrycks2020measuring}, ARC-Easy \cite{clark2018arc}, BoolQ \cite{clark2019boolq}, HellaSwag \cite{zellers2019hellaswag} \\
Seeds (TinyLlama IID) & \{42, 123, 456\} \\
Seeds (LLaMA-3.2-3B) & \{42, 123, 456\} \\
ReverseAdaptive defaults & \texttt{warmup\_rounds}=5, $\tau{=}0.01$, \texttt{stability\_threshold}=1.1, \texttt{transition\_rounds}=0 (instantaneous) \\
\bottomrule
\end{tabular}
\end{table*}

\textbf{Baselines.} FFA-LoRA \cite{sun2024ffalora} and FedIT \cite{zhang2024federatedgpt} are reimplemented within this codebase rather than taken from the original authors' released code, which places all five protocols on one training loop and one byte-accounting path; their absolute results may differ from those of the original implementations. Following Sun et al., the FFA-LoRA implementation freezes $A$ at initialization: client-side $A$ parameters are set non-trainable before local optimization begins and are excluded from the optimizer parameter group, so clients train only $B$ from the first round. Clients transmit $B$ only, a measured 3{,}244{,}032 bytes per client per direction against 9{,}011{,}200 for full $A{+}B$ transmission. The former is $(32768 + 4096) \times 22$ parameters at 4 bytes each, exactly the B-only payload, and could not arise if $A$ were being transmitted. FedIT applies FedAvg independently to $A$ and $B$ and transmits full state in both directions, so its communication volume matches FLoRA's by construction.

\textbf{Prompt format.} Examples are rendered as \texttt{\#\#\# Instruction:} followed by the instruction text, then \texttt{\#\#\# Response:} followed by the output, truncated and padded to 256 tokens. Alpaca's optional \texttt{input} field, populated in roughly 40\% of examples, is not included in training or in evaluation. Dolly-15k differs in training: when its \texttt{context} field is non-empty, a \texttt{\#\#\# Context:} block carrying it is inserted between instruction and response, which applies to 913 of the 3000 training examples. The held-out evaluation omits that block for every example, including the 130 of 500 held-out examples that carry a context, so Dolly-15k held-out losses are measured on a format partly unseen in training. Every method is scored on the same format, so the between-method gaps on Dolly-15k are unaffected; the Dolly-15k levels are not comparable to a context-aware evaluation. Loss is computed over the full formatted sequence including prompt tokens, with padding masked to $-100$. This differs from the two-template formatting used in the original Alpaca release and shifts absolute loss values; because all methods are trained and evaluated identically, comparisons between methods are unaffected. Note that if the tokenizer aliases \texttt{pad\_token} to \texttt{eos\_token}, the end-of-sequence token is masked along with padding; this is consistent across all runs but is relevant to anyone recomputing the base-model reference values of Section~\ref{sec:heldout}.

\textbf{Backend provenance.} Communication totals are protocol-deterministic given the switch round; the backends agree on every IID run at $\tau$=0.01 and differ by one switch round only near a switching boundary (Appendix~\ref{app:crossbackend}). Loss values differ across backends: Appendix~\ref{app:crossbackend} reports cross-backend agreement within $0.01$ for IID settings, which exceeds some of the between-method differences reported below. Comparisons here are therefore made within a backend, and tables are labeled with their corpus. The five-method frontier (Table~\ref{tab:frontier}), the Dolly-15k replication (Table~\ref{tab:dolly}), and all statistical tests use the CUDA corpus, since the FFA-LoRA, FedIT, and Dolly runs exist only there. The threshold ablation (Section~\ref{sec:ablation}) and the LLaMA-3.2-3B results use MPS. Communication values are comparable across tables; loss values are comparable only within a backend. All 34 runs have a counterpart on the second backend, and Appendix~\ref{app:crossbackend} reports the comparison.

\subsection{Held-Out Instruction-Following Metric}
\label{sec:heldout}
Final training loss measures fit to the federated objective, and the four zero-shot benchmarks do not discriminate between base and fine-tuned checkpoints at this scale (Section~\ref{sec:baseregression}). Quality is therefore also reported as held-out instruction-following loss. From each dataset the 500-example slice \texttt{train[3000:3500]} is reserved, disjoint from the 3000 examples partitioned across clients, and mean token-level cross-entropy is computed under the prompt format described in Section~\ref{sec:setup}. Each adapted checkpoint is scored against the unadapted base model, and the reported quantity is $\Delta\ell_{\mathrm{held}} = \ell_{\mathrm{tuned}} - \ell_{\mathrm{base}}$, so more negative values indicate better instruction following. Base-model reference values are 1.9352 (perplexity 6.9254) on Alpaca and 2.2608 (perplexity 9.5905) on Dolly-15k. Because the base values differ by dataset, held-out deltas are comparable within a dataset but not across datasets; the cross-dataset comparison in Section~\ref{sec:dolly} therefore compares between-method gaps rather than levels.

\subsection{The Communication-Quality Frontier}
\label{sec:frontier}
\begin{figure}[tbp]
\centering
\includegraphics[width=\columnwidth]{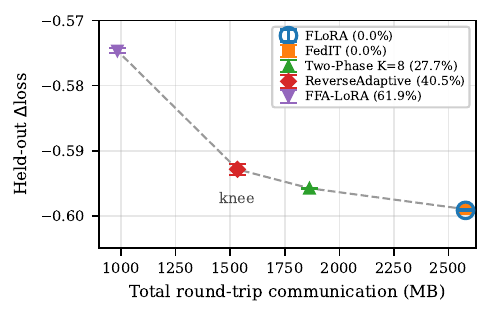}
\caption{Measured communication-quality frontier on Alpaca-3k, TinyLlama-1.1B, IID, CUDA corpus. Five protocols, three from prior work, at four distinct operating points; error bars are one sample standard deviation over three seeds, and communication is identical across seeds. Held-out $\Delta$loss is tuned minus base on the 500-example slice defined in Section~\ref{sec:heldout}; more negative is better. FLoRA (open circle) and FedIT (filled square) coincide at lower right. The segment from ReverseAdaptive to FFA-LoRA is markedly steeper than the segment preceding it, which Section~\ref{sec:measuredbytes} quantifies}
\label{fig:frontier}
\end{figure}

Table~\ref{tab:frontier} and Fig.~\ref{fig:frontier} place five federated LoRA protocols on a single measured communication-quality frontier for TinyLlama-1.1B \cite{zhang2024tinyllama} under IID partitioning. The five methods occupy four distinct operating points. FLoRA \cite{wang2024flora} and FedIT \cite{zhang2024federatedgpt} coincide at 2578.13 MB, since both transmit full $A$ and $B$ in both directions every round. Two-Phase $K{=}8$ reaches 1863.13 MB (27.73\% savings), ReverseAdaptive at $\tau{=}0.01$ reaches 1533.13 MB (40.53\%), and FFA-LoRA \cite{sun2024ffalora} reaches 983.13 MB (61.87\%). Across the four distinct points communication falls and both quality metrics degrade in the same order, so the frontier is monotone on each metric independently rather than only on the metric used to construct it.

\begin{table*}[tbp]
\centering
\footnotesize
\setlength{\tabcolsep}{6pt}
\caption{Measured communication-quality frontier. TinyLlama-1.1B, Alpaca-3k, IID, CUDA corpus, three seeds, mean $\pm$ one sample standard deviation. Communication is protocol-deterministic and identical across seeds. Held-out $\Delta$loss is tuned minus base on the 500-example slice of Section~\ref{sec:heldout}; more negative is better. Differences quoted in the text are computed from unrounded values and may differ from the displayed cells in the final digit. $p$-values are paired $t$-tests against FLoRA on final loss; the complete set of paired tests on both metrics is in Appendix~\ref{app:stats}. Methods: FLoRA \cite{wang2024flora}, FedIT \cite{zhang2024federatedgpt}, FFA-LoRA \cite{sun2024ffalora}. The MB column is total round-trip communication over 15 rounds}
\label{tab:frontier}
\begin{tabular}{@{}lrrrrr@{}}
\toprule
Method & MB & Savings & Final loss & Held-out $\Delta$ & $p$ vs.\ FLoRA \\
\midrule
FLoRA                                 & 2578.13 & --      & $1.2608 \pm 0.0005$ & $-0.5992 \pm 0.0001$ & -- \\
FedIT                                 & 2578.13 & 0.00\%  & $1.2602 \pm 0.0016$ & $-0.5990 \pm 0.0002$ & $0.588$ \\
Two-Phase $K{=}8$                   & 1863.13 & 27.73\% & $1.2705 \pm 0.0006$ & $-0.5958 \pm {<}0.0001$ & $4.2 \times 10^{-5}$ \\
ReverseAdaptive ($\tau{=}0.01$)     & 1533.13 & 40.53\% & $1.2749 \pm 0.0006$ & $-0.5929 \pm 0.0008$ & $1.1 \times 10^{-5}$ \\
FFA-LoRA                             & 983.13 & 61.87\% & $1.3031 \pm 0.0012$ & $-0.5746 \pm 0.0004$ & $1.7 \times 10^{-4}$ \\
\bottomrule
\end{tabular}
\end{table*}

Three features of Table~\ref{tab:frontier} bear on the interpretation. The two quality metrics agree on the ordering of every pair they can resolve: across the four distinct operating points the ranking is identical on final training loss and on held-out instruction-following loss. The exception is the FLoRA and FedIT pair, which shares an operating point, where FedIT is nominally better on final loss and FLoRA nominally better on held-out loss; neither difference is statistically distinguishable ($p{=}0.588$ and $p{=}0.425$). The shape of the frontier is not an artifact of the metric chosen to construct it.

ReverseAdaptive outperforms FFA-LoRA by 0.0282 on final loss and 0.0182 on held-out loss, against per-method seed standard deviations of at most 0.0016 and 0.0008 respectively. Both protocols end training with $A$ frozen and only $B$ aggregated, and they differ chiefly in whether $A$ was learned first; ReverseAdaptive also evaluates its aggregated $B$ with a frozen $A$ that differs slightly from the one its clients trained against (Section~\ref{sec:implementation}). Learning $A$ through a FLoRA phase before freezing it produces materially better adapters than freezing $A$ at initialization, at a cost of 550 MB in this configuration.

FedIT lands on FLoRA's communication volume exactly, to the byte, and within 0.0006 on both quality metrics. It contributes no new operating point, which is the intended result: FedIT is an independently implemented full-state protocol, so its coincidence with FLoRA is evidence that the byte accounting measures the protocol rather than an implementation artifact.

Loss differences between methods are resolved across three seeds: the 0.0141 gap between FLoRA and ReverseAdaptive at $\tau{=}0.01$ is more than twenty times the per-seed standard deviation. Paired $t$-tests on final loss give $p < 10^{-3}$ for every comparison against FLoRA except FedIT, which is correctly indistinguishable. These tests have limited power at three seeds (Appendix~\ref{app:stats}), so effect sizes should be read alongside the $p$-values, and no multiplicity correction is applied in Table~\ref{tab:frontier}; Appendix~\ref{app:stats} reports the complete set of paired tests on both metrics with the correction stated.

This frontier sweeps through the fixed-$K$ Two-Phase operating points: on the MPS corpus, $\tau{=}0.002$ lands on Two-Phase $K{=}8$ communication and $\tau{=}0.001$ on $K{=}10$. Because total communication is a step function of the discrete switch round, where each round of delay changes the total by 110.0\,MB, this correspondence is not backend-invariant; Appendix~\ref{app:crossbackend} reports that both tight-$\tau$ settings fire one round earlier on CUDA. The robust claim is that the adaptive method traverses the same frontier as the fixed-$K$ family without requiring $K$ to be specified, rather than that it reproduces particular $K$ values byte for byte.

\subsection{Replication on Dolly-15k}
\label{sec:dolly}
Table~\ref{tab:dolly} repeats the three-method comparison on Databricks Dolly-15k \cite{dolly2023}, an instruction dataset with a different length distribution and a broader task mix than Alpaca. Communication is identical to the Alpaca corpus, since the protocols are dataset-independent and ReverseAdaptive fired at round 6 on every IID seed of both datasets. The quantity of interest is therefore the quality cost of switching rather than the absolute loss.

\begin{table*}[tbp]
\centering
\footnotesize
\setlength{\tabcolsep}{6pt}
\caption{Dolly-15k replication. TinyLlama-1.1B, IID, CUDA corpus, three seeds, mean $\pm$ one sample standard deviation. Gap is the held-out $\Delta$loss difference against the same dataset's FLoRA baseline, which is the quantity comparable across datasets; held-out levels are not, since base-model losses differ (1.9352 on Alpaca, 2.2608 on Dolly-15k). Differences quoted in the text are computed from unrounded values and may differ from the displayed cells in the final digit. Method: FLoRA \cite{wang2024flora}. Two-Phase and ReverseAdaptive are defined in Section~\ref{sec:method}. The MB column is total round-trip communication over 15 rounds}
\label{tab:dolly}
\begin{tabular}{@{}lrrrrr@{}}
\toprule
Method & MB & Final loss & Held-out $\Delta$ & Gap (Dolly) & Gap (Alpaca) \\
\midrule
FLoRA                           & 2578.13 & $1.6544 \pm 0.0021$ & $-0.5330 \pm 0.0010$ & -- & -- \\
Two-Phase $K{=}8$               & 1863.13 & $1.6643 \pm 0.0021$ & $-0.5287 \pm 0.0010$ & 0.0043 & 0.0034 \\
ReverseAdaptive ($\tau{=}0.01$) & 1533.13 & $1.6686 \pm 0.0022$ & $-0.5268 \pm 0.0005$ & 0.0061 & 0.0063 \\
\bottomrule
\end{tabular}
\end{table*}

Held-out deltas are not comparable across datasets, since the base model scores 1.9352 on the Alpaca slice and 2.2608 on the Dolly slice. Table~\ref{tab:dolly} reports the gap to each dataset's own FLoRA baseline, which is comparable. ReverseAdaptive costs 0.0063 held-out loss on Alpaca and 0.0061 on Dolly, a shift of 0.0002, or 3.0\% of the gap's own magnitude. The quality cost of the adaptive transition is stable across these two datasets.

Two-Phase $K{=}8$ costs 0.0034 on Alpaca and 0.0043 on Dolly, a shift of 25.5\%. The claim supported here is limited to ReverseAdaptive's own cost being stable, not to a property of the protocol family.

A non-IID Dolly run at $\alpha{=}0.5$ switched at round 6 on all three seeds and reached the same 1533.13 MB, matching the IID behavior on both datasets. No Dolly FLoRA baseline was run under heterogeneity, so no quality cost is reported for that configuration.

\subsection{Convergence and Switching Behavior}
\begin{figure*}[tbp]
\centering
\includegraphics[width=\textwidth]{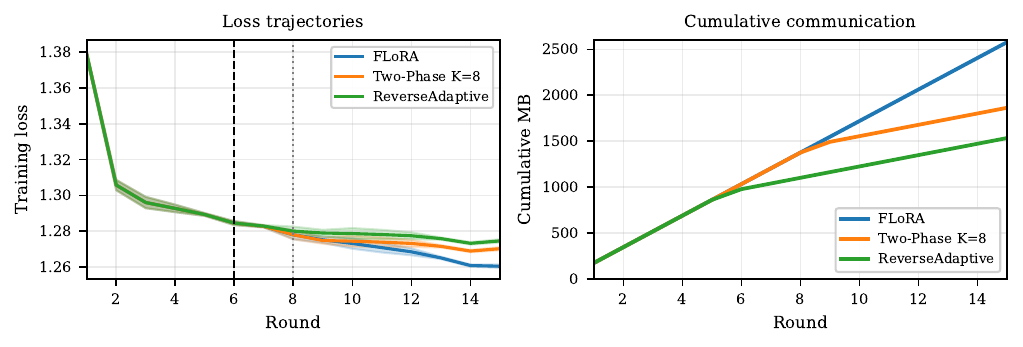}
\caption{Left: training loss trajectories on TinyLlama-1.1B IID, MPS corpus (mean $\pm$ 1 standard deviation over three seeds). Vertical lines mark ReverseAdaptive's switch (round 6) and Two-Phase $K{=}8$'s phase boundary (round 8; the transition round, with full upload, is round 9). Right: cumulative communication on TinyLlama-1.1B (seed 42, deterministic across seeds). Slope changes mark phase transitions; look-ahead download accounting moves the broadcast slope one round before the upload slope}
\label{fig:convergence}
\label{fig:cumcomm}
\end{figure*}
Fig.~\ref{fig:convergence} shows mean $\pm$ std loss trajectories (left) and cumulative communication (right) across three seeds for FLoRA, Two-Phase $K{=}8$, and ReverseAdaptive. All three methods converge similarly through the FLoRA phase. Phase transitions are visible as slight slope changes but do not destabilize training. The final ${\sim}0.01$ loss gap accumulates gradually after the switch rather than appearing as a sharp discontinuity. FLoRA stays linear at 85.94 MB per round per direction throughout. Two-Phase $K{=}8$ maintains the same slope through round 9, then reduces as B-only transmission activates. ReverseAdaptive splits off at round 6 (at TinyLlama-1.1B scale), producing the steepest savings trajectory.

The no-switch sanity baseline (ReverseAdaptive with \mbox{\texttt{switch\_threshold}}\,{=}\,{-}1.0, \mbox{\texttt{warmup\_rounds}}\,{=}\,999) produces final loss 1.2594342812 on seed 42 under the MPS corpus, matching the corresponding FLoRA run to 10 decimal places. The CUDA per-seed values in Appendix~\ref{app:perseed} differ, as expected across backends; the guarantee is a within-backend statement. Upload and download match to the byte across all 15 rounds.
Disabling the switch requires a negative threshold rather than a small positive one, for reasons set out in Appendix~\ref{app:noswitch}.

\subsection{Downstream Evaluation}
Tables~\ref{tab:downstream_tiny}--\ref{tab:downstream_llama} present downstream results on MMLU \cite{hendrycks2020measuring}, ARC-Easy \cite{clark2018arc}, BoolQ \cite{clark2019boolq}, and HellaSwag \cite{zellers2019hellaswag}. MMLU is excluded from TinyLlama-1.1B comparisons: base accuracy of 0.250 on 4-way multiple choice indicates chance-level performance, making the benchmark uninformative at 1.1B scale.

At TinyLlama-1.1B (Table~\ref{tab:downstream_tiny}), all federated methods cluster within 1.0 percentage points of each other across the three informative benchmarks and regress modestly relative to the base model (Section~\ref{sec:baseregression}).

\begin{table*}[tbp]
\centering
\footnotesize
\setlength{\tabcolsep}{6pt}
\caption{Downstream zero-shot accuracy, TinyLlama-1.1B, single training seed (42), MPS corpus, 500 examples per benchmark. MMLU omitted: base accuracy is at chance for 4-way multiple choice}
\label{tab:downstream_tiny}
\begin{tabular}{@{}lrrrr@{}}
\toprule
Method & ARC-Easy & BoolQ & HellaSwag & $\Delta$ vs.\ base (BoolQ) \\
\midrule
Base (no adapter) & 0.274 & 0.626 & 0.448 & -- \\
FLoRA             & 0.246 & 0.562 & 0.420 & $-6.4$ pp \\
Two-Phase $K{=}8$ & 0.248 & 0.560 & 0.420 & $-6.6$ pp \\
ReverseAdaptive   & 0.252 & 0.552 & 0.420 & $-7.4$ pp \\
\midrule
Spread (max--min) & 0.006 & 0.010 & 0.000 & -- \\
\bottomrule
\end{tabular}
\end{table*}

At LLaMA-3.2-3B scale, the three federated methods are indistinguishable on downstream benchmarks within seed variation. Mean accuracies across three seeds differ by at most 0.5 percentage points on any benchmark, and the per-method error bars overlap on every benchmark. Relative to the base model they do not regress as at 1.1B (Section~\ref{sec:baseregression}): the LLaMA base achieves MMLU 0.544, ARC-Easy 0.832, BoolQ 0.706, and HellaSwag 0.540, and every federated method matches or exceeds the first three while sitting 0.1 to 0.5 percentage points below the base on HellaSwag.

\begin{table*}[bp]
\centering
\footnotesize
\setlength{\tabcolsep}{6pt}
\caption{Downstream zero-shot accuracy, LLaMA-3.2-3B. Mean $\pm$ 1 standard deviation across three random seeds (42, 123, 456). 500 examples per benchmark. Cross-method spread is at most 0.5 percentage points on any benchmark and the per-method error bars overlap}
\label{tab:downstream_llama}
\begin{tabular}{@{}lrrrr@{}}
\toprule
Method & MMLU & ARC-Easy & BoolQ & HellaSwag \\
\midrule
Base (no adapter) & 0.544 & 0.832 & 0.706 & 0.540 \\
FLoRA             & $0.558 \pm 0.007$ & $0.836 \pm 0.007$ & $0.736 \pm 0.007$ & $0.535 \pm 0.003$ \\
Two-Phase $K{=}8$ & $0.558 \pm 0.002$ & $0.837 \pm 0.008$ & $0.738 \pm 0.010$ & $0.539 \pm 0.004$ \\
ReverseAdaptive   & $0.557 \pm 0.003$ & $0.839 \pm 0.008$ & $0.739 \pm 0.004$ & $0.538 \pm 0.004$ \\
\bottomrule
\end{tabular}
\end{table*}

\subsection{Threshold Ablation}
\label{sec:ablation}
\begin{figure}[tbp]
\centering
\includegraphics[width=\columnwidth]{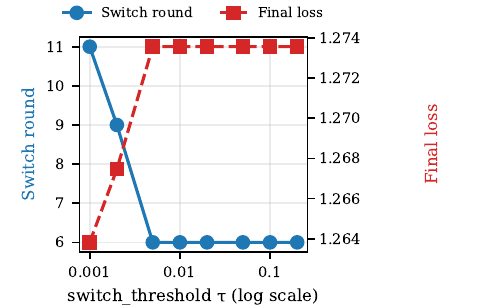}
\caption{ReverseAdaptive threshold sensitivity (TinyLlama IID, seed 42, MPS corpus). Loss values are not comparable with the CUDA results in Table~\ref{tab:frontier}, and the $\tau$=0.001 and $\tau$=0.002 settings switch one round earlier on CUDA (Appendix~\ref{app:crossbackend})}
\label{fig:ablation}
\end{figure}

Table~\ref{tab:ablation} and Fig.~\ref{fig:ablation} characterize how $\tau$ controls the savings-quality tradeoff. Since $\tau$ bounds a relative rather than an absolute loss improvement, the values below are fractions of the previous round's loss: $\tau{=}0.01$ requires the loss to fall by at least 1\% in a round for the FLoRA phase to continue. The method has a sensitive regime for $\tau \leq 0.002$: at $\tau{=}0.002$ the switch occurs at round 9, matching Two-Phase $K{=}8$; at $\tau{=}0.001$ the switch occurs at round 11, matching Two-Phase $K{=}10$. For $\tau \geq 0.005$ the method saturates: the switch fires at round 6, the first round after warmup. All rows for $\tau \in \{0.005, 0.010, 0.020, 0.050, 0.100, 0.200\}$ produce identical results.

A practitioner seeking maximum savings can set $\tau$ anywhere in $[0.005, 0.2]$ without precision loss. The $\tau{=}0.001$ and $\tau{=}0.002$ operating points show the switch round responding to the loss trajectory.

\begin{table*}[tbp]
\centering
\footnotesize
\setlength{\tabcolsep}{6pt}
\caption{ReverseAdaptive threshold sensitivity (TinyLlama IID, seed 42, MPS corpus)}
\label{tab:ablation}
\begin{tabular}{@{}lrrrr@{}}
\toprule
$\tau$ & Switch round & Final loss & Total comm.\ (MB) & Savings \\
\midrule
0.001 & 11 & 1.2638 & 2083.13 & 19.20\% \\
0.002 & 9  & 1.2675 & 1863.13 & 27.73\% \\
0.005--0.2 & 6  & 1.2736 & 1533.13 & 40.53\% \\
\bottomrule
\end{tabular}
\end{table*}

\subsection{Scale Validation}
\begin{figure*}[tbp]
\centering
\includegraphics[width=\textwidth]{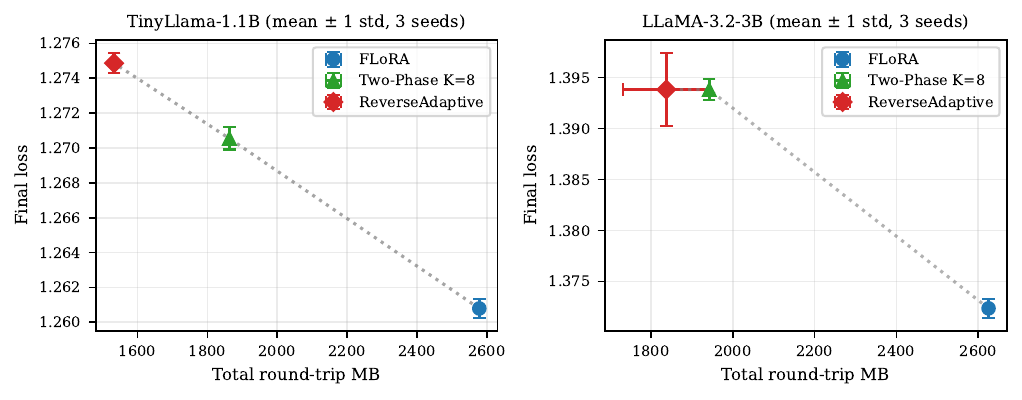}
\caption{Scale validation. Left: TinyLlama-1.1B, CUDA corpus (mean $\pm$ 1 standard deviation over three seeds). Right: LLaMA-3.2-3B, MPS corpus (mean $\pm$ 1 standard deviation over three seeds). Both panels show the three federated methods on the same axes. Dotted polyline connects the three method points on each panel. The Pareto structure (downward slope from FLoRA to ReverseAdaptive) is preserved across scales. Y-axis ranges differ between panels because absolute training loss is model-dependent; the relative Pareto shape is the invariant being validated. ReverseAdaptive's larger error bar at LLaMA-3.2-3B reflects across-seed variance in switch round (7, 8, 9)}
\label{fig:scale}
\end{figure*}

Table~\ref{tab:llama_iid} and Fig.~\ref{fig:scale} confirm that the Pareto structure observed at TinyLlama-1.1B is preserved at LLaMA-3.2-3B.

Three random seeds (42, 123, 456) were run for each of FLoRA, Two-Phase $K{=}8$, and ReverseAdaptive, producing nine total runs. Final-round results (mean $\pm$ standard deviation): FLoRA reaches loss $1.3724 \pm 0.0009$ with 2625.0 MB total communication. Two-Phase $K{=}8$ reaches loss $1.3938 \pm 0.0011$ with 1942.5 MB (savings 26.0\%). ReverseAdaptive reaches loss $1.3938 \pm 0.0036$ with $1837.5 \pm 105.0$ MB (savings $30.0 \pm 4.0$\%). The protocol-deterministic communication for FLoRA and Two-Phase $K{=}8$ produces zero seed-to-seed variance; ReverseAdaptive's variance reflects across-seed differences in switch round (7, 8, 9 across the three seeds, mean $8.0 \pm 1.0$).

\begin{table*}[tbp]
\centering
\footnotesize
\setlength{\tabcolsep}{6pt}
\caption{LLaMA-3.2-3B IID results. ReverseAdaptive uses $\tau{=}0.01$. The MB column is total round-trip communication over 15 rounds and savings are against FLoRA. Mean $\pm$ 1 standard deviation across three random seeds (42, 123, 456). ReverseAdaptive switch rounds: 7, 8, 9. FLoRA and Two-Phase $K{=}8$ communication is protocol-deterministic; ReverseAdaptive communication varies with switch round}
\label{tab:llama_iid}
\begin{tabular}{@{}lrrr@{}}
\toprule
Method & Final loss & MB & Savings \\
\midrule
FLoRA & $1.3724 \pm 0.0009$ & 2625.0 & 0\% \\
Two-Phase $K{=}8$ & $1.3938 \pm 0.0011$ & 1942.5 & 26.0\% \\
ReverseAdaptive & $1.3938 \pm 0.0036$ & $1837.5 \pm 105.0$ & $30.0 \pm 4.0$\% \\
\bottomrule
\end{tabular}
\end{table*}

The communication savings are slightly lower at 3B than at 1.1B: 26.0\% vs.\ 27.7\% for Two-Phase $K{=}8$, and 30.0\% vs.\ 40.5\% for ReverseAdaptive. The B-only fraction of the full state is 40\% at LLaMA-3.2-3B (vs.\ 36\% for TinyLlama) due to different GQA configurations, meaning each B-only round saves a smaller fraction of the full payload at 3B scale.

At LLaMA-3.2-3B, ReverseAdaptive and Two-Phase $K{=}8$ reach final losses differing by $1.25 \times 10^{-5}$, roughly 350 times smaller than the 0.0043 separating them at TinyLlama-1.1B. A paired $t$-test over three seeds gives $p = 0.997$ with a 95\% confidence interval of $[{-}0.0114, 0.0114]$. That interval contains the 1.1B effect size, so the test cannot distinguish no difference from a difference as large as the one observed at 1.1B, and the result is reported as a statement about effect size rather than as evidence of equivalence. ReverseAdaptive also saves 4.0 percentage points more mean communication than the fixed $K{=}8$ configuration (30.0\% against 26.0\%), because two of its three seeds switch before round 9, at rounds 7 and 8. The same threshold $\tau{=}0.01$ used at TinyLlama-1.1B produced this behavior with no scale-specific tuning.

\section{Discussion}
\label{sec:discussion}

\subsection{Implications for Federated LLM Deployment}
In mobile or Internet of Things (IoT) federated settings where uplink bandwidth is scarce and metered, cutting round-trip communication by 30 to 40\% can change the economic viability of federated fine-tuning. At TinyLlama-1.1B the 40.5\% reduction costs 0.0063 in held-out instruction-following loss. Whether that trade is acceptable is a deployment decision rather than a universal one, and the frontier in Section~\ref{sec:frontier} is intended to let a practitioner make it explicitly: the same measurements show that accepting a further 21.3 points of savings costs roughly five times more quality per point. Kairouz et al. \cite{kairouz2021advances} identify communication as a fundamental open problem in federated learning at scale, and measured byte-level reporting is a prerequisite for addressing it.

\subsection{Choosing an Operating Point}
\label{sec:choosing}
The Two-Phase protocol requires a practitioner to select $K$ before training begins, without knowing when the loss plateau will occur for a given model-dataset combination. This is a hyperparameter that must be tuned per dataset, per model, and potentially per client population: precisely the kind of manual configuration that slows deployment in production federated systems. ReverseAdaptive replaces it by monitoring the signal that $K$ was meant to approximate.

The criterion transfers across model scales because it is scale-free. Since $\tau$ bounds a relative loss improvement, the same numerical value encodes the same stopping condition at any absolute loss level, so it carries across models whose losses differ in magnitude; an absolute improvement threshold would not, since a fixed loss delta that plateaus one model's training would fire immediately or never on another. Using the same $\tau{=}0.01$, ReverseAdaptive fires at round 6 on TinyLlama-1.1B and at a mean of round $8.0 \pm 1.0$ on LLaMA-3.2-3B, with no scale-specific retuning. At $\tau{=}0.01$ the IID switch rounds at both scales match exactly across the MPS and CUDA backends (see Appendix~\ref{app:crossbackend}), so the cross-scale comparison does not rest on corpora that disagree about when the switch occurred.

Two details refine this result. At TinyLlama scale and $\tau{=}0.01$ the switch fires at round 6 in every MPS run across all three partition settings and in every Dolly-15k run, and round 6 is the first round after the warmup period $W{=}5$; on CUDA one $\alpha{=}0.1$ partition fires at round 7 instead, as Appendix~\ref{app:crossbackend} reports. At $\tau{=}0.01$ and this scale the switch therefore fires as early as the warmup allows, which is the knee of the frontier; smaller thresholds move the switch along the frontier, with $\tau \leq 0.002$ yielding switch rounds of 9 and 11 (Section~\ref{sec:ablation}). Across seeds in IID training, switch round varies only at LLaMA-3.2-3B, where three runs give rounds 7, 8, and 9; those runs also differ in code revision (Section~\ref{sec:limitations}), so the variation across those three runs cannot be attributed to training dynamics alone. Two model sizes and two datasets do not establish that $\tau{=}0.01$ is a universal default; they establish that a single value carried across the two scales tested without retuning.

At 3B the adaptive and fixed-$K$ configurations reach final losses differing by $1.25 \times 10^{-5}$, roughly 350 times smaller than the 0.0043 that separates them at 1.1B. A paired $t$-test over three seeds gives $p = 0.997$. This is reported as a statement about effect size rather than as evidence of equivalence: with three seeds the test has almost no power to detect a difference of the magnitude observed at 1.1B, so a large $p$-value cannot establish that none exists.

Where ReverseAdaptive sits on the frontier matters more than which fixed $K$ it happens to match. Moving from FLoRA to ReverseAdaptive buys 40.5 percentage points of communication savings at a held-out cost of 0.0063, or $1.56 \times 10^{-4}$ per point. Moving onward to FFA-LoRA buys a further 21.3 points at a cost of 0.0182, or $8.54 \times 10^{-4}$ per point, 5.5 times more expensive per point saved. The frontier has a knee and ReverseAdaptive sits at it. A practitioner who takes the first tranche of savings and declines the second is making the trade the measured data supports, and the value of the adaptive rule is that it locates that point without $K$ being guessed in advance.

\needspace{4\baselineskip}
\subsection{Measured Bytes and Theoretical Parameter Counts}
\label{sec:measuredbytes}
Federated LoRA work commonly reports communication savings as parameter-count ratios \cite{zhang2024federatedgpt,sun2024ffalora,wang2024flora,bai2024flexlora}, conflating how many parameters exist with how many bytes a protocol transmits. The practice is not universal: FedSRD \cite{yan2026fedsrd} and FLoRIST \cite{ramesh2026florist} both report measured per-round communication in megabytes. Aggregate megabyte totals, however, do not separate two quantities that govern switching protocols and that a parameter ratio cannot express.

Architectural choices such as GQA fix the B-only fraction in ways a parameter ratio does not anticipate. TinyLlama-1.1B's GQA makes B-only transmission save exactly 36\% rather than the 50\% a non-GQA count suggests, and LLaMA-3.2-3B's different GQA configuration yields exactly 40\%. A theoretical 50\% savings claim made for one model family does not transfer to another without architectural accounting.

The transition round carries a cost that parameter-ratio analyses omit. At the FLoRA-to-FFA-LoRA boundary the upload is full while the broadcast is already B-only, so exactly one directional payload is upgraded from B-only to full. On TinyLlama this costs 55.0 MB, and the figure does not depend on the switch round: totals that ignore the transition would be 928.13, 1478.13, and 1808.13 MB for switch rounds 1, 6, and 9, against measured totals of 983.13, 1533.13, and 1863.13. What varies with the number of remaining rounds is the penalty's share of the total, not its size. A protocol upgrading both directions at the boundary would pay 110.0 MB; the look-ahead broadcast halves it.

Total communication across the entire corpus is one closed-form expression in one discrete parameter. For a switching protocol with switch round $s$ over $N$ rounds, uploads are full for rounds $1 \ldots s$ and broadcasts for rounds $1 \ldots s{-}1$, so $2s-1$ directional payloads are full and the total is $55(2s-1) + 928.125$ MB on TinyLlama and $52.5(2s-1) + 1050$ MB on LLaMA-3.2-3B, where $928.125$ and the non-switching endpoint $2578.125$ are exact. FFA-LoRA is $s{=}1$, ReverseAdaptive $s{=}6$, Two-Phase $K{=}8$ is $s{=}9$, and Two-Phase $K{=}10$ is $s{=}11$; every measured total reported at full ten-client participation lands on that lattice to the byte; the two reduced-participation runs in Appendix~\ref{app:perseed} scale off the lattice in proportion to the active client count. The $2s-1$ arises from the harvest rule rather than from the protocol: an implementation that froze the server's existing global $A$ would upgrade no directional payload at the boundary and give $110(s-1) + 928.125$ instead. A non-switching protocol has all $2N$ directional payloads full and is not a point on the lattice: FLoRA is the separate endpoint at $55 \times 30 + 928.125 = 2578.125$ MB. The lattice spacing of 110 MB per round is also why no continuous $\tau$ reproduces a fixed-$K$ total robustly, as Section~\ref{sec:frontier} notes.

This implementation charges FFA-LoRA slightly more than its design requires. FFA-LoRA freezes $A$ at initialization, so every client's $A$ equals the shared initial value at every round and never needs transmitting. The measured 983.13 MB nonetheless includes the 55.0 MB seeding upload, because the server harvests $A$ from the first client's uploaded state rather than deriving it from the initialization (Section~\ref{sec:implementation}). An implementation faithful to the design reaches the floor of 928.13 MB, raising FFA-LoRA's savings from 61.87\% to 64.00\%, widening the ReverseAdaptive-to-FFA-LoRA segment from 21.3 to 23.5 percentage points, and moving the marginal cost ratio from 5.5 to 5.0. The knee conclusion is unchanged. The same harvest rule governs Two-Phase and ReverseAdaptive. For Two-Phase the harvested $A$ is the round-$K$ global $A$ the server already holds, so freezing that value directly would leave training unchanged and remove the 55.0\,MB seeding upload. For ReverseAdaptive the harvested $A$ carries one epoch of local adaptation, so freezing the held global $A$ instead would change the adapter as well as the bytes, and is untested (Section~\ref{sec:limitations}).

\subsection{Zero-Shot Benchmark Behavior and the Training Objective}
\label{sec:baseregression}
Federated Alpaca-3k \cite{taori2023alpaca} instruction tuning regresses TinyLlama-1.1B \cite{zhang2024tinyllama} on all three informative zero-shot benchmarks: ARC-Easy by 2.2 to 2.8 percentage points depending on method, BoolQ by 6.4 to 7.4, and HellaSwag by 2.8 uniformly. At LLaMA-3.2-3B \cite{dubey2024llama3} the same procedure is flat-to-positive on MMLU, ARC-Easy, and BoolQ and within 0.5 percentage points of the base on HellaSwag. The cross-scale contrast is consistent with a model-capacity explanation: TinyLlama-1.1B may lack the parameter budget to absorb instruction tuning without distributional drift on these benchmarks, though a single dataset and two scales cannot rule out other explanations.

The same procedure improves substantially on the objective it optimizes. On the held-out Alpaca slice the base model scores 1.9352 (perplexity 6.9254) and every fine-tuned checkpoint scores between 1.3349 and 1.3608 (perplexity 3.80 to 3.90). The regression is a divergence between the training objective and these benchmarks rather than a training failure, and it is why quality is reported here as held-out instruction-following loss. The held-out slice is drawn from the same distribution as the training data, so it measures instruction following on unseen examples and is not evidence of generalization to unrelated tasks.

The regression is independent of aggregation method. Cross-method spread at TinyLlama scale is 0.006 on ARC-Easy, 0.010 on BoolQ, and 0.000 on HellaSwag, so the zero-shot benchmarks do not separate protocols that the held-out metric separates at $p < 0.05$. These accuracies are single-seed (seed 42, MPS corpus), unlike the three-seed CUDA results in Table~\ref{tab:frontier}, so the spreads carry no variance estimate and indicate insensitivity rather than establishing equivalence.

The BoolQ result admits a specific account. The base model's accuracy of 0.626 closely matches BoolQ's natural yes-class rate of approximately 0.62, suggesting the base exploits a yes-biased prior. Instruction tuning attenuates that prior, which is why BoolQ moves furthest of the three.

\section{Limitations}
\label{sec:limitations}

The experiments use two instruction-following datasets (Alpaca-3k \cite{taori2023alpaca} and Dolly-15k \cite{dolly2023}) and a single task type. Generalization to classification tasks, longer training runs, or generation-quality metrics is not characterized. Quality is measured as held-out loss on examples drawn from the same distribution as the training data, which captures instruction following on unseen examples rather than transfer to unrelated tasks.

FFA-LoRA and FedIT were evaluated on Alpaca only, so Table~\ref{tab:dolly} compares three protocols where Table~\ref{tab:frontier} compares five. The cross-dataset result therefore speaks to the cost of the adaptive transition rather than to the stability of the whole frontier across datasets.

The study covers only Llama-style decoder-only architectures with LoRA targeting \mbox{\texttt{q\_proj} and \texttt{v\_proj}}. Other architectures, encoder-decoder models, larger target sets, or heterogeneous rank configurations may produce different B-only fractions and different savings profiles. The reported percentages are specific to this target set and to the grouped-query attention configurations of the two models tested.

The simulation assumes full client participation in every round. Real federated deployments typically involve partial participation, where only a fraction of clients respond per round. The effect of partial participation on transition-round timing and on ReverseAdaptive's plateau detection is not studied here.

Non-IID partitions are constructed by Dirichlet skew over instruction-length buckets rather than over semantic labels; Alpaca carries no class labels, and Dolly-15k's category field was not used. This induces heterogeneity in task form. Heterogeneity in task semantics, which is what semantic-class partitioning models in classification settings, is untested.

No real-world network conditions are simulated. Byte counts are measured from the tensors actually transmitted, but network latency, packet loss, and bandwidth constraints are not modeled, so wall-clock communication time is not predicted.

No differential privacy (DP) analysis is included. FFA-LoRA \cite{sun2024ffalora} was originally motivated in part by DP-friendliness; the transition-round dynamics of the bidirectional B-only protocol may affect DP accounting in non-obvious ways that warrant separate study.

Fine-tuned checkpoints score below the base model on zero-shot benchmarks at TinyLlama scale (Section~\ref{sec:baseregression}), so conclusions about relative protocol quality rest on held-out instruction-following loss.

At the transition round the server freezes $A$ from the first client's uploaded state. For ReverseAdaptive this value differs from the $A$ against which clients train $B$ after the switch, and the effect of that mismatch on its reported quality is not measured (Section~\ref{sec:implementation}). Freezing the global $A$ the server already holds would remove the seeding upload and is untested, so the 55.0\,MB transition cost and ReverseAdaptive's quality figures are properties of this implementation rather than of phase-switching protocols in general.

Runs were produced across more than one code revision, some including local modifications that were not committed at run time. The primary CUDA corpus spans two revisions, and the FLoRA and FedIT rows of Table~\ref{tab:frontier} sit on either side of that boundary while remaining statistically indistinguishable, which indicates that the combined algorithmic and revision difference across that boundary is roughly 0.0006 in final loss. The LLaMA-3.2-3B results are not single-revision: the three seeds were produced at three revisions, so the variability reported there combines seed and environment variation.

LLaMA-3.2-3B experiments use three random seeds with downstream evaluation per checkpoint. Seed-to-seed evaluation noise within a single checkpoint is not separately characterized. At TinyLlama-1.1B, downstream evaluation is single-seed per checkpoint while training uses three seeds.

Future work should address partial participation, non-Llama architectures, semantic heterogeneity, and real network conditions to validate deployment claims more broadly. Composing the bidirectional B-only protocol with gradient compression techniques \cite{alistarh2017qsgd,lin2017deep} for savings beyond the B-only floor is a natural extension. Per-layer adaptive phase-switching, where different transformer layers switch at different rounds based on their individual loss-plateau signals, is another direction.

\section{Conclusion}
\label{sec:conclusion}

This paper places five federated LoRA protocols, three from prior work, on a single measured communication-quality frontier and shows that the frontier has a knee. Savings beyond the knee cost roughly five times more held-out quality per point than savings up to it, and ReverseAdaptive locates that point without a phase boundary being specified in advance. Learning $A$ through a FLoRA phase before freezing it also produces materially better adapters than freezing $A$ at initialization: 0.0182 in held-out loss against FFA-LoRA, roughly twenty times the largest per-method seed standard deviation, at a cost of 550\,MB.

The frontier rests on byte-level measurement rather than parameter-count ratios. Tracking per-round upload and download megabytes, including at the asymmetric transition boundary, isolates two quantities parameter-count accounting cannot express: the grouped-query attention B-only fraction, 36\% on TinyLlama-1.1B and 40\% on LLaMA-3.2-3B rather than 50\%, and a transition cost of 55.0\,MB per switch in this implementation, independent of when the switch occurs.

Two checks bound the measurement. An independently implemented full-state protocol lands on the same total to the byte, and a closed form in the switch round reproduces every measured total at full participation. The 34-run core of the MPS corpus was re-executed on a second hardware backend; of the 31 with a valid reference, 27 agree within tolerance, and every communication difference between backends is exactly one step of that lattice.

The same threshold transfers across model scales without retuning, firing at round 6 on TinyLlama-1.1B and at a mean of round 8 on LLaMA-3.2-3B, and the quality cost of the transition is stable across the two datasets tested.

As open-weight LLMs continue to grow in capability and federated deployments expand into privacy-sensitive domains, the communication bottleneck identified by Kairouz et al. \cite{kairouz2021advances} will only intensify. Reporting what protocols actually transmit, rather than what a parameter count predicts, is a prerequisite for comparing them.

\backmatter

\bmhead{Acknowledgements}

The Alpaca and Dolly-15k datasets are publicly available from their respective sources.

The author used Claude (Anthropic) for grammar and language editing of the manuscript and as a coding assistant for the released codebase. All technical content, analysis, and interpretations are the author's own.

{\tolerance=2000\emergencystretch=1.5em
\bibliography{main}

%% BioMed_Central_Bib_Style_v1.01

\begin{thebibliography}{26}
% BibTex style file: bmc-mathphys.bst (version 2.1), 2014-07-24
\ifx \bisbn   \undefined \def \bisbn  #1{ISBN #1}\fi
\ifx \binits  \undefined \def \binits#1{#1}\fi
\ifx \bauthor  \undefined \def \bauthor#1{#1}\fi
\ifx \batitle  \undefined \def \batitle#1{#1}\fi
\ifx \bjtitle  \undefined \def \bjtitle#1{#1}\fi
\ifx \bvolume  \undefined \def \bvolume#1{\textbf{#1}}\fi
\ifx \byear  \undefined \def \byear#1{#1}\fi
\ifx \bissue  \undefined \def \bissue#1{#1}\fi
\ifx \bfpage  \undefined \def \bfpage#1{#1}\fi
\ifx \blpage  \undefined \def \blpage #1{#1}\fi
\ifx \burl  \undefined \def \burl#1{\textsf{#1}}\fi
\ifx \doiurl  \undefined \def \doiurl#1{\url{https://doi.org/#1}}\fi
\ifx \betal  \undefined \def \betal{\textit{et al.}}\fi
\ifx \binstitute  \undefined \def \binstitute#1{#1}\fi
\ifx \binstitutionaled  \undefined \def \binstitutionaled#1{#1}\fi
\ifx \bctitle  \undefined \def \bctitle#1{#1}\fi
\ifx \beditor  \undefined \def \beditor#1{#1}\fi
\ifx \bpublisher  \undefined \def \bpublisher#1{#1}\fi
\ifx \bbtitle  \undefined \def \bbtitle#1{#1}\fi
\ifx \bedition  \undefined \def \bedition#1{#1}\fi
\ifx \bseriesno  \undefined \def \bseriesno#1{#1}\fi
\ifx \blocation  \undefined \def \blocation#1{#1}\fi
\ifx \bsertitle  \undefined \def \bsertitle#1{#1}\fi
\ifx \bsnm \undefined \def \bsnm#1{#1}\fi
\ifx \bsuffix \undefined \def \bsuffix#1{#1}\fi
\ifx \bparticle \undefined \def \bparticle#1{#1}\fi
\ifx \barticle \undefined \def \barticle#1{#1}\fi
\bibcommenthead
\ifx \bconfdate \undefined \def \bconfdate #1{#1}\fi
\ifx \botherref \undefined \def \botherref #1{#1}\fi
\ifx \url \undefined \def \url#1{\textsf{#1}}\fi
\ifx \bchapter \undefined \def \bchapter#1{#1}\fi
\ifx \bbook \undefined \def \bbook#1{#1}\fi
\ifx \bcomment \undefined \def \bcomment#1{#1}\fi
\ifx \oauthor \undefined \def \oauthor#1{#1}\fi
\ifx \citeauthoryear \undefined \def \citeauthoryear#1{#1}\fi
\ifx \endbibitem  \undefined \def \endbibitem {}\fi
\ifx \bconflocation  \undefined \def \bconflocation#1{#1}\fi
\ifx \arxivurl  \undefined \def \arxivurl#1{\textsf{#1}}\fi
\csname PreBibitemsHook\endcsname

%%% 1
\bibitem[\protect\citeauthoryear{Kairouz et~al.}{2021}]{kairouz2021advances}
\begin{barticle}
\bauthor{\bsnm{Kairouz}, \binits{P.}},
\bauthor{\bsnm{McMahan}, \binits{H.B.}},
\bauthor{\bsnm{Avent}, \binits{B.}},
\bauthor{\bsnm{Bellet}, \binits{A.}},
\bauthor{\bsnm{Bennis}, \binits{M.}},
\bauthor{\bsnm{Bhagoji}, \binits{A.N.}}, \betal:
\batitle{Advances and open problems in federated learning}.
\bjtitle{Found. Trends Mach. Learn.}
\bvolume{14}(\bissue{1--2}),
\bfpage{1}--\blpage{210}
(\byear{2021})
\doiurl{10.1561/2200000083}
\end{barticle}
\endbibitem

%%% 2
\bibitem[\protect\citeauthoryear{Hu et~al.}{2022}]{hu2022lora}
\begin{bchapter}
\bauthor{\bsnm{Hu}, \binits{E.J.}},
\bauthor{\bsnm{Shen}, \binits{Y.}},
\bauthor{\bsnm{Wallis}, \binits{P.}},
\bauthor{\bsnm{Allen-Zhu}, \binits{Z.}},
\bauthor{\bsnm{Li}, \binits{Y.}},
\bauthor{\bsnm{Wang}, \binits{S.}},
\bauthor{\bsnm{Wang}, \binits{L.}},
\bauthor{\bsnm{Chen}, \binits{W.}}:
\bctitle{{LoRA}: Low-rank adaptation of large language models}.
In: \bbtitle{International Conference on Learning Representations (ICLR)}
(\byear{2022})
\end{bchapter}
\endbibitem

%%% 3
\bibitem[\protect\citeauthoryear{Zhang et~al.}{2024}]{zhang2024federatedgpt}
\begin{bchapter}
\bauthor{\bsnm{Zhang}, \binits{J.}},
\bauthor{\bsnm{Vahidian}, \binits{S.}},
\bauthor{\bsnm{Kuo}, \binits{M.}},
\bauthor{\bsnm{Li}, \binits{C.}},
\bauthor{\bsnm{Zhang}, \binits{R.}},
\bauthor{\bsnm{Yu}, \binits{T.}},
\bauthor{\bsnm{Wang}, \binits{G.}},
\bauthor{\bsnm{Chen}, \binits{Y.}}:
\bctitle{Towards building the {FederatedGPT}: Federated instruction tuning}.
In: \bbtitle{IEEE International Conference on Acoustics, Speech and Signal
  Processing (ICASSP)},
pp. \bfpage{6915}--\blpage{6919}
(\byear{2024}).
\doiurl{10.1109/ICASSP48485.2024.10447454}
\end{bchapter}
\endbibitem

%%% 4
\bibitem[\protect\citeauthoryear{McMahan
  et~al.}{2017}]{mcmahan2017communication}
\begin{bchapter}
\bauthor{\bsnm{McMahan}, \binits{B.}},
\bauthor{\bsnm{Moore}, \binits{E.}},
\bauthor{\bsnm{Ramage}, \binits{D.}},
\bauthor{\bsnm{Hampson}, \binits{S.}},
\bauthor{\bsnm{Arcas}, \binits{B.}}:
\bctitle{Communication-efficient learning of deep networks from decentralized
  data}.
In: \bbtitle{International Conference on Artificial Intelligence and Statistics
  (AISTATS)}
(\byear{2017})
\end{bchapter}
\endbibitem

%%% 5
\bibitem[\protect\citeauthoryear{Sun et~al.}{2024}]{sun2024ffalora}
\begin{bchapter}
\bauthor{\bsnm{Sun}, \binits{Y.}},
\bauthor{\bsnm{Li}, \binits{Z.}},
\bauthor{\bsnm{Li}, \binits{Y.}},
\bauthor{\bsnm{Ding}, \binits{B.}}:
\bctitle{Improving {LoRA} in privacy-preserving federated learning}.
In: \bbtitle{International Conference on Learning Representations (ICLR)}
(\byear{2024}).
\bcomment{arXiv:2403.12313}
\end{bchapter}
\endbibitem

%%% 6
\bibitem[\protect\citeauthoryear{Wang et~al.}{2024}]{wang2024flora}
\begin{bchapter}
\bauthor{\bsnm{Wang}, \binits{Z.}},
\bauthor{\bsnm{Shen}, \binits{Z.}},
\bauthor{\bsnm{He}, \binits{Y.}},
\bauthor{\bsnm{Sun}, \binits{G.}},
\bauthor{\bsnm{Wang}, \binits{H.}},
\bauthor{\bsnm{Lyu}, \binits{L.}},
\bauthor{\bsnm{Li}, \binits{A.}}:
\bctitle{{FLoRA}: Federated fine-tuning large language models with
  heterogeneous low-rank adaptations}.
In: \bbtitle{Advances in Neural Information Processing Systems (NeurIPS)},
pp. \bfpage{22513}--\blpage{22533}
(\byear{2024}).
\bcomment{arXiv:2409.05976}
\end{bchapter}
\endbibitem

%%% 7
\bibitem[\protect\citeauthoryear{Bai et~al.}{2024}]{bai2024flexlora}
\begin{bchapter}
\bauthor{\bsnm{Bai}, \binits{J.}},
\bauthor{\bsnm{Chen}, \binits{D.}},
\bauthor{\bsnm{Qian}, \binits{B.}},
\bauthor{\bsnm{Yao}, \binits{L.}},
\bauthor{\bsnm{Li}, \binits{Y.}}:
\bctitle{Federated fine-tuning of large language models under heterogeneous
  tasks and client resources}.
In: \bbtitle{Advances in Neural Information Processing Systems (NeurIPS)},
pp. \bfpage{14457}--\blpage{14483}
(\byear{2024}).
\bcomment{arXiv:2402.11505}
\end{bchapter}
\endbibitem

%%% 8
\bibitem[\protect\citeauthoryear{Yan et~al.}{2026}]{yan2026fedsrd}
\begin{bchapter}
\bauthor{\bsnm{Yan}, \binits{G.}},
\bauthor{\bsnm{Xie}, \binits{L.}},
\bauthor{\bsnm{Shen}, \binits{Q.}},
\bauthor{\bsnm{Fang}, \binits{Y.}},
\bauthor{\bsnm{Wu}, \binits{Z.}}:
\bctitle{{FedSRD}: Sparsify-reconstruct-decompose for communication-efficient
  federated large language models fine-tuning}.
In: \bbtitle{Proceedings of the ACM Web Conference 2026 (WWW '26)},
pp. \bfpage{5087}--\blpage{5098}
(\byear{2026}).
\doiurl{10.1145/3774904.3792144}
\end{bchapter}
\endbibitem

%%% 9
\bibitem[\protect\citeauthoryear{Ramesh and Dass}{2026}]{ramesh2026florist}
\begin{bchapter}
\bauthor{\bsnm{Ramesh}, \binits{H.}},
\bauthor{\bsnm{Dass}, \binits{J.}}:
\bctitle{{FLoRIST}: Singular value thresholding for efficient and accurate
  federated fine-tuning of large language models}.
In: \bbtitle{Proceedings of the Ninth Conference on Machine Learning and
  Systems (MLSys)}
(\byear{2026}).
\bcomment{arXiv:2506.09199v2}
\end{bchapter}
\endbibitem

%%% 10
\bibitem[\protect\citeauthoryear{Guo et~al.}{2025}]{guo2025fedsalora}
\begin{bchapter}
\bauthor{\bsnm{Guo}, \binits{P.}},
\bauthor{\bsnm{Zeng}, \binits{S.}},
\bauthor{\bsnm{Wang}, \binits{Y.}},
\bauthor{\bsnm{Fan}, \binits{H.}},
\bauthor{\bsnm{Wang}, \binits{F.}},
\bauthor{\bsnm{Qu}, \binits{L.}}:
\bctitle{Selective aggregation for low-rank adaptation in federated learning}.
In: \bbtitle{International Conference on Learning Representations (ICLR)}
(\byear{2025}).
\bcomment{arXiv:2410.01463}
\end{bchapter}
\endbibitem

%%% 11
\bibitem[\protect\citeauthoryear{Alistarh et~al.}{2017}]{alistarh2017qsgd}
\begin{bchapter}
\bauthor{\bsnm{Alistarh}, \binits{D.}},
\bauthor{\bsnm{Grubic}, \binits{D.}},
\bauthor{\bsnm{Li}, \binits{J.}},
\bauthor{\bsnm{Tomioka}, \binits{R.}},
\bauthor{\bsnm{Vojnovic}, \binits{M.}}:
\bctitle{{QSGD}: Communication-efficient {SGD} via gradient quantization and
  encoding}.
In: \bbtitle{Advances in Neural Information Processing Systems (NeurIPS)}
(\byear{2017})
\end{bchapter}
\endbibitem

%%% 12
\bibitem[\protect\citeauthoryear{Lin et~al.}{2018}]{lin2017deep}
\begin{bchapter}
\bauthor{\bsnm{Lin}, \binits{Y.}},
\bauthor{\bsnm{Han}, \binits{S.}},
\bauthor{\bsnm{Mao}, \binits{H.}},
\bauthor{\bsnm{Wang}, \binits{Y.}},
\bauthor{\bsnm{Dally}, \binits{W.J.}}:
\bctitle{Deep gradient compression: Reducing the communication bandwidth for
  distributed training}.
In: \bbtitle{International Conference on Learning Representations (ICLR)}
(\byear{2018})
\end{bchapter}
\endbibitem

%%% 13
\bibitem[\protect\citeauthoryear{Reisizadeh
  et~al.}{2020}]{reisizadeh2020fedpaq}
\begin{bchapter}
\bauthor{\bsnm{Reisizadeh}, \binits{A.}},
\bauthor{\bsnm{Mokhtari}, \binits{A.}},
\bauthor{\bsnm{Hassani}, \binits{H.}},
\bauthor{\bsnm{Jadbabaie}, \binits{A.}},
\bauthor{\bsnm{Pedarsani}, \binits{R.}}:
\bctitle{{FedPAQ}: A communication-efficient federated learning method with
  periodic averaging and quantization}.
In: \bbtitle{International Conference on Artificial Intelligence and Statistics
  (AISTATS)}
(\byear{2020})
\end{bchapter}
\endbibitem

%%% 14
\bibitem[\protect\citeauthoryear{Li et~al.}{2020}]{li2020fedprox}
\begin{bchapter}
\bauthor{\bsnm{Li}, \binits{T.}},
\bauthor{\bsnm{Sahu}, \binits{A.K.}},
\bauthor{\bsnm{Zaheer}, \binits{M.}},
\bauthor{\bsnm{Sanjabi}, \binits{M.}},
\bauthor{\bsnm{Talwalkar}, \binits{A.}},
\bauthor{\bsnm{Smith}, \binits{V.}}:
\bctitle{Federated optimization in heterogeneous networks}.
In: \bbtitle{Proceedings of Machine Learning and Systems (MLSys)}
(\byear{2020})
\end{bchapter}
\endbibitem

%%% 15
\bibitem[\protect\citeauthoryear{Wang et~al.}{2020}]{wang2020fednova}
\begin{bchapter}
\bauthor{\bsnm{Wang}, \binits{J.}},
\bauthor{\bsnm{Liu}, \binits{Q.}},
\bauthor{\bsnm{Liang}, \binits{H.}},
\bauthor{\bsnm{Joshi}, \binits{G.}},
\bauthor{\bsnm{Poor}, \binits{H.V.}}:
\bctitle{Tackling the objective inconsistency problem in heterogeneous
  federated optimization}.
In: \bbtitle{Advances in Neural Information Processing Systems (NeurIPS)}
(\byear{2020})
\end{bchapter}
\endbibitem

%%% 16
\bibitem[\protect\citeauthoryear{Reddi et~al.}{2021}]{reddi2020adaptive}
\begin{bchapter}
\bauthor{\bsnm{Reddi}, \binits{S.}},
\bauthor{\bsnm{Charles}, \binits{Z.}},
\bauthor{\bsnm{Zaheer}, \binits{M.}},
\bauthor{\bsnm{Garrett}, \binits{Z.}},
\bauthor{\bsnm{Rush}, \binits{K.}},
\bauthor{\bsnm{Kone\v{c}n\'{y}}, \binits{J.}},
\bauthor{\bsnm{Kumar}, \binits{S.}},
\bauthor{\bsnm{McMahan}, \binits{H.B.}}:
\bctitle{Adaptive federated optimization}.
In: \bbtitle{International Conference on Learning Representations (ICLR)}
(\byear{2021})
\end{bchapter}
\endbibitem

%%% 17
\bibitem[\protect\citeauthoryear{Loshchilov and
  Hutter}{2019}]{loshchilov2019decoupled}
\begin{bchapter}
\bauthor{\bsnm{Loshchilov}, \binits{I.}},
\bauthor{\bsnm{Hutter}, \binits{F.}}:
\bctitle{Decoupled weight decay regularization}.
In: \bbtitle{International Conference on Learning Representations (ICLR)}
(\byear{2019})
\end{bchapter}
\endbibitem

%%% 18
\bibitem[\protect\citeauthoryear{Zhang et~al.}{2024}]{zhang2024tinyllama}
\begin{botherref}
\oauthor{\bsnm{Zhang}, \binits{P.}},
\oauthor{\bsnm{Zeng}, \binits{G.}},
\oauthor{\bsnm{Wang}, \binits{T.}},
\oauthor{\bsnm{Lu}, \binits{W.}}:
{TinyLlama}: An open-source small language model.
arXiv preprint
(2024).
arXiv:2401.02385
\end{botherref}
\endbibitem

%%% 19
\bibitem[\protect\citeauthoryear{Dubey et~al.}{2024}]{dubey2024llama3}
\begin{botherref}
\oauthor{\bsnm{Dubey}, \binits{A.}},
\oauthor{\bsnm{Jauhri}, \binits{A.}},
\oauthor{\bsnm{Pandey}, \binits{A.}},
\oauthor{\bsnm{Kadian}, \binits{A.}},
\oauthor{\bsnm{Al-Dahle}, \binits{A.}}, et al.:
The {Llama} 3 herd of models.
arXiv preprint
(2024).
arXiv:2407.21783
\end{botherref}
\endbibitem

%%% 20
\bibitem[\protect\citeauthoryear{Taori et~al.}{2023}]{taori2023alpaca}
\begin{botherref}
\oauthor{\bsnm{Taori}, \binits{R.}},
\oauthor{\bsnm{Gulrajani}, \binits{I.}},
\oauthor{\bsnm{Zhang}, \binits{T.}},
\oauthor{\bsnm{Dubois}, \binits{Y.}},
\oauthor{\bsnm{Li}, \binits{X.}},
\oauthor{\bsnm{Guestrin}, \binits{C.}},
\oauthor{\bsnm{Liang}, \binits{P.}},
\oauthor{\bsnm{Hashimoto}, \binits{T.B.}}:
Alpaca: A Strong, Replicable Instruction-Following Model.
Stanford CRFM Blog.
Accessed 2 October 2026
(2023).
\url{https://crfm.stanford.edu/2023/03/13/alpaca.html}
\end{botherref}
\endbibitem

%%% 21
\bibitem[\protect\citeauthoryear{Conover et~al.}{2023}]{dolly2023}
\begin{botherref}
\oauthor{\bsnm{Conover}, \binits{M.}},
\oauthor{\bsnm{Hayes}, \binits{M.}},
\oauthor{\bsnm{Mathur}, \binits{A.}},
\oauthor{\bsnm{Xie}, \binits{J.}},
\oauthor{\bsnm{Wan}, \binits{J.}},
\oauthor{\bsnm{Shah}, \binits{S.}},
\oauthor{\bsnm{Ghodsi}, \binits{A.}},
\oauthor{\bsnm{Wendell}, \binits{P.}},
\oauthor{\bsnm{Zaharia}, \binits{M.}},
\oauthor{\bsnm{Xin}, \binits{R.}}:
Free Dolly: Introducing the World's First Truly Open Instruction-Tuned {LLM}.
Databricks blog.
Accessed 2 October 2026
(2023).
\url{https://www.databricks.com/blog/2023/04/12/dolly-first-open-commercially-viable-instruction-tuned-llm}
\end{botherref}
\endbibitem

%%% 22
\bibitem[\protect\citeauthoryear{Hendrycks
  et~al.}{2021}]{hendrycks2020measuring}
\begin{bchapter}
\bauthor{\bsnm{Hendrycks}, \binits{D.}},
\bauthor{\bsnm{Burns}, \binits{C.}},
\bauthor{\bsnm{Basart}, \binits{S.}},
\bauthor{\bsnm{Zou}, \binits{A.}},
\bauthor{\bsnm{Mazeika}, \binits{M.}},
\bauthor{\bsnm{Song}, \binits{D.}},
\bauthor{\bsnm{Steinhardt}, \binits{J.}}:
\bctitle{Measuring massive multitask language understanding}.
In: \bbtitle{International Conference on Learning Representations (ICLR)}
(\byear{2021})
\end{bchapter}
\endbibitem

%%% 23
\bibitem[\protect\citeauthoryear{Clark et~al.}{2018}]{clark2018arc}
\begin{botherref}
\oauthor{\bsnm{Clark}, \binits{P.}},
\oauthor{\bsnm{Cowhey}, \binits{I.}},
\oauthor{\bsnm{Etzioni}, \binits{O.}},
\oauthor{\bsnm{Khot}, \binits{T.}},
\oauthor{\bsnm{Sabharwal}, \binits{A.}},
\oauthor{\bsnm{Schoenick}, \binits{C.}},
\oauthor{\bsnm{Tafjord}, \binits{O.}}:
Think you have solved question answering? {Try ARC}, the {AI2} reasoning
  challenge.
arXiv preprint
(2018).
arXiv:1803.05457
\end{botherref}
\endbibitem

%%% 24
\bibitem[\protect\citeauthoryear{Clark et~al.}{2019}]{clark2019boolq}
\begin{bchapter}
\bauthor{\bsnm{Clark}, \binits{C.}},
\bauthor{\bsnm{Lee}, \binits{K.}},
\bauthor{\bsnm{Chang}, \binits{M.-W.}},
\bauthor{\bsnm{Kwiatkowski}, \binits{T.}},
\bauthor{\bsnm{Collins}, \binits{M.}},
\bauthor{\bsnm{Toutanova}, \binits{K.}}:
\bctitle{{BoolQ}: Exploring the surprising difficulty of natural yes/no
  questions}.
In: \bbtitle{Proceedings of NAACL-HLT},
pp. \bfpage{2924}--\blpage{2936}
(\byear{2019}).
\doiurl{10.18653/v1/N19-1300}
\end{bchapter}
\endbibitem

%%% 25
\bibitem[\protect\citeauthoryear{Zellers et~al.}{2019}]{zellers2019hellaswag}
\begin{bchapter}
\bauthor{\bsnm{Zellers}, \binits{R.}},
\bauthor{\bsnm{Holtzman}, \binits{A.}},
\bauthor{\bsnm{Bisk}, \binits{Y.}},
\bauthor{\bsnm{Farhadi}, \binits{A.}},
\bauthor{\bsnm{Choi}, \binits{Y.}}:
\bctitle{{HellaSwag}: Can a machine really finish your sentence?}
In: \bbtitle{Proceedings of the Association for Computational Linguistics
  (ACL)},
pp. \bfpage{4791}--\blpage{4800}
(\byear{2019}).
\doiurl{10.18653/v1/P19-1472}
\end{bchapter}
\endbibitem

%%% 26
\bibitem[\protect\citeauthoryear{Franklin}{2026}]{franklin2026fedloraprotocols}
\begin{botherref}
\oauthor{\bsnm{Franklin}, \binits{J.A.}}:
fedlora-protocols: Communication-efficient federated {LoRA} fine-tuning,
  release v1.0.1.
Zenodo
(2026).
\doiurl{10.5281/zenodo.23111915}
\end{botherref}
\endbibitem

\end{thebibliography}
}

\section*{Statements and Declarations}

\begin{itemize}
\item \textbf{Funding:} The author received no funding for this work.
\item \textbf{Competing interests:} The author declares no competing interests.
\item \textbf{Ethics approval and consent to participate:} Not applicable.
\item \textbf{Consent for publication:} Not applicable.
\item \textbf{Data availability:} Analysis CSVs, run records, and scripts to regenerate tables and figures are available in the public repository at \url{https://github.com/jerryadamsfranklin/fedlora-protocols} and archived as release v1.0.1 at \doiurl{10.5281/zenodo.23111915}~\cite{franklin2026fedloraprotocols}. Final adapter states for the three seed-42 TinyLlama-1.1B runs evaluated in Table~\ref{tab:downstream_tiny} are included; other adapter checkpoints are not. A verification script in the repository recomputes the paper's principal quantitative claims from the released records.
\item \textbf{Materials availability:} Not applicable.
\item \textbf{Code availability:} Source code is available in the same repository.
\item \textbf{Author contributions:} The author performed all work associated with this study.
\end{itemize}

\begin{appendices}
% sn-jnl already sets Alph-section table/figure labels inside appendices;
% attach those counters to section so each appendix restarts at 1 (B1, C1, D1).
\makeatletter
\@addtoreset{table}{section}%
\@addtoreset{figure}{section}%
\makeatother
\section{Experimental Details}
\label{app:hyper}
All experiments use AdamW \cite{loshchilov2019decoupled} with learning rate $10^{-4}$, $\beta_1{=}0.9$, $\beta_2{=}0.999$, and weight decay 0.01 (the PyTorch default). No learning rate schedule is applied. Gradients are clipped to a maximum $\ell_2$ norm of 1.0 at every optimizer step. Maximum sequence length is 256 tokens. See Table~\ref{tab:setup} for the complete configuration including ReverseAdaptive defaults.

\subsection{Reproducibility Checklist}
\label{app:repro}
\textbf{Code:} Full source code is publicly released, including all configuration YAML files, training scripts, evaluation scripts, and scripts to reproduce every figure and table from the analysis CSVs.

\textbf{Verification:} A script in the repository (\texttt{scripts/verify\_numbers.py}) recomputes the paper's principal quantitative claims from the released analysis files and run records with 179 automated assertions.

\textbf{Seeds:} The IID seeds are listed in Table~\ref{tab:setup}. The $\alpha{=}0.1$ partition-sensitivity runs additionally use seeds 789 and 1000, which appear only in the partition-sensitivity results of Appendix~\ref{app:perseed}, alongside the other per-seed results.

\textbf{Hardware:} The MPS corpus was executed on a single Apple M4 Pro workstation with 48\,GB unified memory; a core of 34 of its runs, approximately 285 hours of compute, was re-executed on commercially rented NVIDIA hardware, an RTX 4090 for TinyLlama-1.1B and an A100 40\,GB for LLaMA-3.2-3B; the comparison is in Appendix~\ref{app:crossbackend}. The Dolly-15k replication and the FFA-LoRA and FedIT baselines were produced on the rented hardware only.

\textbf{Data:} Alpaca \cite{taori2023alpaca} via \texttt{tatsu-lab/alpaca} on HuggingFace, 3000-sample subset, and Dolly-15k \cite{dolly2023} via \texttt{databricks/databricks-dolly-15k}, matched subset. Adapter checkpoints are not released, except the final adapter states of the three seed-42 TinyLlama-1.1B runs evaluated in Table~\ref{tab:downstream_tiny}.

\subsection{Disabling the Switch}
\label{app:noswitch}
Running ReverseAdaptive with switching disabled reproduces the matched FLoRA \cite{wang2024flora} run to 10 decimal places at every round, establishing that the wrapper is a zero-cost modification when unused: it rules out numerical artifacts from the wrapper's presence, not merely from its activation. The comparison is within a single backend and a single execution environment. Agreement across hardware is weaker than bit-identity and is reported in Appendix~\ref{app:crossbackend}.

Disabling the switch requires care. Setting $\tau$ to a small positive value does not achieve it: any round in which the training loss fails to improve satisfies $\rho_r < \tau$, so the criterion fires. On the FLoRA trajectory underlying the threshold ablation the loss rises once, at round 11, so $\tau{=}0$ still triggers a switch at that round, and suppressing the switch entirely requires $\tau$ below the most negative round-over-round relative change observed, approximately ${-}0.0009$ here. The sanity baseline accordingly uses $\tau{=}{-}1.0$ together with a warmup exceeding the round budget. A practitioner wanting the wrapper present but inert should disable it by one of those means rather than by choosing a conservative threshold.

\needspace{3\baselineskip}
\section{Per-Seed Full Results}
\label{app:perseed}
Tables~\ref{tab:b1} to~\ref{tab:b4} give per-seed results and are labeled with their corpus, which follows the corresponding table in the main text. The TinyLlama-1.1B method comparisons use the CUDA corpus, since the FFA-LoRA, FedIT, and Dolly-15k runs exist only there; the LLaMA-3.2-3B results use MPS, matching Table~\ref{tab:llama_iid}; and the partition-sensitivity results retain the larger MPS seed set, where five seeds at $\alpha{=}0.1$ provide better evidence than three.

\begin{table*}[tbp]
\centering
\footnotesize
\scriptsize
\renewcommand{\arraystretch}{0.78}
\setlength{\tabcolsep}{6pt}
\caption{Per-seed TinyLlama-1.1B IID results, CUDA corpus, backing Table~\ref{tab:frontier}. Communication is deterministic across seeds. For Two-Phase the Switch/boundary column shows the predetermined phase boundary $K$, not an adaptive switch event; for FFA-LoRA, $A$ is frozen from initialization (shown as ``init''); for ReverseAdaptive it shows the round at which the plateau signal triggered}
\label{tab:b1}
\begin{tabular}{@{}llrrr@{}}
\toprule
Method & Seed & Final loss & Total MB & Switch / boundary \\
\midrule
FLoRA & 42 & 1.261408 & 2578.13 & -- \\
FLoRA & 123 & 1.260479 & 2578.13 & -- \\
FLoRA & 456 & 1.260500 & 2578.13 & -- \\
FedIT & 42 & 1.260767 & 2578.13 & -- \\
FedIT & 123 & 1.258420 & 2578.13 & -- \\
FedIT & 456 & 1.261503 & 2578.13 & -- \\
Two-Phase $K{=}8$ & 42 & 1.271276 & 1863.13 & $K{=}8$ \\
Two-Phase $K{=}8$ & 123 & 1.270130 & 1863.13 & $K{=}8$ \\
Two-Phase $K{=}8$ & 456 & 1.270237 & 1863.13 & $K{=}8$ \\
ReverseAdaptive ($\tau{=}0.01$) & 42 & 1.275514 & 1533.13 & 6 \\
ReverseAdaptive ($\tau{=}0.01$) & 123 & 1.274610 & 1533.13 & 6 \\
ReverseAdaptive ($\tau{=}0.01$) & 456 & 1.274481 & 1533.13 & 6 \\
FFA-LoRA & 42 & 1.303975 & 983.13 & init \\
FFA-LoRA & 123 & 1.301738 & 983.13 & init \\
FFA-LoRA & 456 & 1.303628 & 983.13 & init \\
\bottomrule
\end{tabular}
\end{table*}

At $\alpha{=}0.5$, mean final loss is statistically indistinguishable from IID for every method in the released analysis CSVs, with differences of 0.001 to 0.007 against per-seed standard deviations near 0.02 over three seeds. Heterogeneity manifests instead as sensitivity to the partition draw: seed-to-seed standard deviation rises by roughly an order of magnitude, from $0.5$ to $1.6 \times 10^{-3}$ under IID in Table~\ref{tab:b1} to $18$ to $25 \times 10^{-3}$ at $\alpha{=}0.5$ across the methods in the released CSVs, which is also why the per-seed spread at $\alpha{=}0.1$ in Table~\ref{tab:b2} is wide.

\begin{table*}[tbp]
\centering
\footnotesize
\setlength{\tabcolsep}{6pt}
\caption{Non-IID results, MPS corpus, five seeds at $\alpha{=}0.1$. This table documents partition sensitivity rather than comparing methods, so the larger MPS seed set is retained; Table~\ref{tab:b1} follows the CUDA corpus because it backs a method comparison. Seeds 42 and 1000 at $\alpha{=}0.1$ had 9 and 8 active clients due to empty Dirichlet partitions, which scales communication to $9/10$ and $8/10$ of the full value (1379.81 and 1226.50 MB against 1533.13). Cross-backend verification (Appendix~\ref{app:crossbackend}) covers seeds 42, 123, and 456; seeds 789 and 1000 were executed on MPS only}
\label{tab:b2}
\begin{tabular}{@{}llrrrr@{}}
\toprule
Setting & Seed & Final loss & Total MB & Switch round & Active clients \\
\midrule
ReverseAdaptive, $\alpha{=}0.5$ & 42 & 1.285933 & 1533.13 & 6 & 10 \\
ReverseAdaptive, $\alpha{=}0.5$ & 123 & 1.282625 & 1533.13 & 6 & 10 \\
ReverseAdaptive, $\alpha{=}0.5$ & 456 & 1.252133 & 1533.13 & 6 & 10 \\
ReverseAdaptive, $\alpha{=}0.1$ & 42 & 1.268353 & 1379.81 & 6 & 9 \\
ReverseAdaptive, $\alpha{=}0.1$ & 123 & 1.288985 & 1533.13 & 6 & 10 \\
ReverseAdaptive, $\alpha{=}0.1$ & 456 & 1.348083 & 1533.13 & 6 & 10 \\
ReverseAdaptive, $\alpha{=}0.1$ & 789 & 1.259200 & 1533.13 & 6 & 10 \\
ReverseAdaptive, $\alpha{=}0.1$ & 1000 & 1.273046 & 1226.50 & 6 & 8 \\
\bottomrule
\end{tabular}
\end{table*}

\begin{table*}[tbp]
\centering
\footnotesize
\setlength{\tabcolsep}{6pt}
\caption{Per-seed LLaMA-3.2-3B IID results, MPS corpus, backing Table~\ref{tab:llama_iid}. ReverseAdaptive uses $\tau{=}0.01$. Communication is deterministic given the switch round; the three ReverseAdaptive seeds occupy three adjacent lattice points}
\label{tab:b3}
\begin{tabular}{@{}llrrr@{}}
\toprule
Method & Seed & Final loss & Total MB & Switch \\
\midrule
FLoRA & 42 & 1.371683 & 2625.0 & -- \\
FLoRA & 123 & 1.373402 & 2625.0 & -- \\
FLoRA & 456 & 1.371998 & 2625.0 & -- \\
Two-Phase $K{=}8$ & 42 & 1.393776 & 1942.5 & $K{=}8$ \\
Two-Phase $K{=}8$ & 123 & 1.394877 & 1942.5 & $K{=}8$ \\
Two-Phase $K{=}8$ & 456 & 1.392772 & 1942.5 & $K{=}8$ \\
ReverseAdaptive & 42 & 1.393831 & 1837.5 & 8 \\
ReverseAdaptive & 123 & 1.390265 & 1942.5 & 9 \\
ReverseAdaptive & 456 & 1.397366 & 1732.5 & 7 \\
\bottomrule
\end{tabular}
\end{table*}

\begin{table*}[tbp]
\centering
\footnotesize
\setlength{\tabcolsep}{6pt}
\caption{Per-seed TinyLlama-1.1B Dolly-15k IID results, CUDA corpus, backing Table~\ref{tab:dolly}. Held-out $\Delta$ is tuned minus base on the Dolly-15k held-out slice; more negative is better}
\label{tab:b4}
\begin{tabular}{@{}llrrr@{}}
\toprule
Method & Seed & Final loss & Held-out $\Delta$ & Total MB \\
\midrule
FLoRA & 42 & 1.656762 & $-0.531848$ & 2578.13 \\
FLoRA & 123 & 1.652934 & $-0.533376$ & 2578.13 \\
FLoRA & 456 & 1.653426 & $-0.533664$ & 2578.13 \\
Two-Phase $K{=}8$ & 42 & 1.666616 & $-0.527571$ & 1863.13 \\
Two-Phase $K{=}8$ & 123 & 1.662685 & $-0.529113$ & 1863.13 \\
Two-Phase $K{=}8$ & 456 & 1.663475 & $-0.529410$ & 1863.13 \\
ReverseAdaptive & 42 & 1.671039 & $-0.526568$ & 1533.13 \\
ReverseAdaptive & 123 & 1.666995 & $-0.527418$ & 1533.13 \\
ReverseAdaptive & 456 & 1.667699 & $-0.526547$ & 1533.13 \\
\bottomrule
\end{tabular}
\end{table*}

\section{Statistical Tests}
\label{app:stats}
Table~\ref{tab:c1} reports paired $t$-tests over three seeds on final-round training loss and on held-out instruction-following loss. TinyLlama-1.1B results use the CUDA corpus, matching Table~\ref{tab:frontier}; LLaMA-3.2-3B results use the MPS corpus.

With three seeds these tests have very low power, so effect sizes should be read alongside the $p$-values. No multiplicity correction is applied in the main results tables. Applying a Bonferroni correction across the six TinyLlama-1.1B Alpaca IID comparisons gives a threshold of $0.0083$, which the ReverseAdaptive vs.\ Two-Phase $K{=}8$ comparison on held-out loss ($p = 0.026$) does not meet. That comparison is secondary under the frontier framing of Section~\ref{sec:discussion}; the load-bearing comparison is ReverseAdaptive vs.\ FFA-LoRA, which holds at $p < 10^{-3}$ on both metrics with a difference roughly twenty times the largest per-method standard deviation.

The LLaMA-3.2-3B comparison between ReverseAdaptive and Two-Phase $K{=}8$ has a 95\% confidence interval of $[{-}0.0114, 0.0114]$, which contains the $0.0043$ difference observed between the same methods at TinyLlama-1.1B. The test therefore cannot distinguish equivalence from a difference of the magnitude seen at the smaller scale.

\begin{table*}[tbp]
\centering
\footnotesize
\setlength{\tabcolsep}{5pt}
\renewcommand{\arraystretch}{0.88}
\caption{Paired $t$-tests, three seeds. TinyLlama-1.1B rows use the CUDA corpus; LLaMA-3.2-3B rows use MPS. Positive $\Delta$ indicates the first method has higher loss. Held-out instruction-following loss was not computed for the LLaMA-3.2-3B corpus; quality at that scale is reported as final training loss and downstream zero-shot accuracy}
\label{tab:c1}
\begin{tabular}{@{}lrrrr@{}}
\toprule
Comparison & $\Delta$ final loss & $p$ & $\Delta$ held-out & $p$ \\
\midrule
\multicolumn{5}{l}{\emph{TinyLlama-1.1B, Alpaca, IID}} \\
Two-Phase $K{=}8$ vs.\ FLoRA & $+0.0098$ & $4.2 \times 10^{-5}$ & $+0.0034$ & $7.9 \times 10^{-5}$ \\
ReverseAdaptive vs.\ FLoRA & $+0.0141$ & $1.1 \times 10^{-5}$ & $+0.0063$ & $6.2 \times 10^{-3}$ \\
ReverseAdaptive vs.\ Two-Phase $K{=}8$ & $+0.0043$ & $3.4 \times 10^{-4}$ & $+0.0029$ & $2.6 \times 10^{-2}$ \\
FFA-LoRA vs.\ ReverseAdaptive & $+0.0282$ & $4.4 \times 10^{-4}$ & $+0.0182$ & $3.8 \times 10^{-4}$ \\
FFA-LoRA vs.\ FLoRA & $+0.0423$ & $1.7 \times 10^{-4}$ & $+0.0245$ & $1.1 \times 10^{-4}$ \\
FedIT vs.\ FLoRA & ${-}0.0006$ & $0.588$ & $+0.0002$ & $0.425$ \\
\midrule
\multicolumn{5}{l}{\emph{TinyLlama-1.1B, Dolly-15k, IID}} \\
Two-Phase $K{=}8$ vs.\ FLoRA & $+0.0099$ & $7.8 \times 10^{-5}$ & $+0.0043$ & $2.4 \times 10^{-6}$ \\
ReverseAdaptive vs.\ FLoRA & $+0.0142$ & $2.5 \times 10^{-5}$ & $+0.0061$ & $7.6 \times 10^{-3}$ \\
\midrule
\multicolumn{5}{l}{\emph{LLaMA-3.2-3B, Alpaca, IID}} \\
Two-Phase $K{=}8$ vs.\ FLoRA & $+0.0214$ & $3.2 \times 10^{-4}$ & -- & -- \\
ReverseAdaptive vs.\ FLoRA & $+0.0215$ & $1.3 \times 10^{-2}$ & -- & -- \\
ReverseAdaptive vs.\ Two-Phase $K{=}8$ & $+0.000012$ & $0.997$ & -- & -- \\
\bottomrule
\end{tabular}
\end{table*}

\section{Cross-Backend Verification}
\label{app:crossbackend}

The primary corpus was produced on the MPS backend and subsequently re-executed on CUDA. Table~\ref{tab:d1} summarizes the comparison; Table~\ref{tab:d2} details every disagreement.

Tolerances are setting-aware. IID runs require per-round training loss to agree within $0.01$ absolute. Non-IID runs allow $0.025$, since heterogeneous partitions amplify per-round variation, but additionally require the switch round to match exactly and permit no more than two consecutive rounds exceeding $0.01$. Communication must agree within $0.01$ in all settings, which is effectively exact since totals are protocol-deterministic.

Of 34 run pairs, 31 have a valid MPS reference. The three Two-Phase $K{=}8$ runs at $\alpha{=}0.5$ do not: their MPS executions predate the bidirectional B-only implementation and never switched, recording 2578.13 MB, the full FLoRA volume. Those are absent references rather than disagreements, and CUDA is the source of truth for that configuration. Of the 31 comparable pairs, 27 agree within tolerance and 4 do not.

\begin{table}[tbp]
\centering
\footnotesize
\renewcommand{\arraystretch}{0.88}
\setlength{\tabcolsep}{6pt}
\caption{Cross-backend verification summary, MPS against CUDA, 34 run pairs}
\label{tab:d1}
\begin{tabular}{@{}lr@{}}
\toprule
Category & Runs \\
\midrule
Agreed within setting-aware tolerance & 27 \\
Disagreed & 4 \\
\midrule
Comparable pairs & 31 \\
No valid MPS reference (CUDA reported) & 3 \\
\midrule
Total & 34 \\
\bottomrule
\end{tabular}
\end{table}

\begin{table}[tbp]
\centering
\footnotesize
\scriptsize
\renewcommand{\arraystretch}{0.85}
\setlength{\tabcolsep}{3pt}
\caption{Cross-backend disagreements. Every communication difference is one 110\,MB lattice step. Run 16 failed only the consecutive-spike rule}
\label{tab:d2}
\begin{tabular}{@{}llrrr@{}}
\toprule
Run & Config. & MPS & CUDA & Note \\
\midrule
11 & $\tau{=}0.001$, IID & 2083.13 & 1973.13 & $s$: 11 to 10 \\
12 & $\tau{=}0.002$, IID & 1863.13 & 1753.13 & $s$: 9 to 8 \\
16 & $\alpha{=}0.1$, 123 & 1533.13 & 1533.13 & spike rule \\
17 & $\alpha{=}0.1$, 456 & 1533.13 & 1643.13 & $s$: 6 to 7 \\
\bottomrule
\end{tabular}
\end{table}

Three of the four disagreements are the same phenomenon: every observed cross-backend communication difference is exactly one 110\,MB switch-round step (Section~\ref{sec:measuredbytes}), so Table~\ref{tab:d2} lists adjacent lattice points. The backends agree on the protocol and disagree only on which round the plateau fires, and only near a boundary: two tight-$\tau$ IID runs ($\tau{=}0.001$, $\tau{=}0.002$) and one $\alpha{=}0.1$ run. Run 16 has no communication or switch-round difference: only the consecutive-spike rule failed.

Run 17 was re-executed on CUDA with partition seed 456 and training seed 999 and again switched at round 7, so the one-round gap is a stable backend/partition property rather than run noise. At $\alpha{=}0.1$, exact switch-round agreement is therefore not guaranteed (one of three CUDA counterparts fires one round later); CUDA is reported as primary. The switch round matched exactly for every IID run at $\tau{=}0.01$ and for all three LLaMA-3.2-3B ReverseAdaptive runs, supporting the cross-scale claim in Section~\ref{sec:choosing}.

\textbf{Byte accounting.} Transport-layer counts use \texttt{numel}$\times$\texttt{element\_size} on tensors actually sent; the FFA-LoRA \texttt{get\_communication\_cost} helper is unused.

\textbf{Code revisions.} Runs span multiple revisions (including uncommitted working-tree changes), so numbers come from released records and the replication here. Byte totals are revision-invariant: FLoRA, Two-Phase $K{=}8$ and ReverseAdaptive each matched under two CUDA revisions at 2578.13, 1863.13 and 1533.13\,MB. Loss carries revision uncertainty; the FLoRA and FedIT rows of Table~\ref{tab:frontier} differ by 0.0006 across that boundary, against a ReverseAdaptive-to-FFA-LoRA gap of 0.0282.
\end{appendices}

\end{document}